\PassOptionsToPackage{hyphens}{url}
\documentclass[11pt]{article}

\usepackage[final]{acl}
\usepackage{fontawesome5}
\usepackage{latexsym}
\usepackage{balance}
\usepackage{xcolor}
\usepackage{times}  
\usepackage{helvet}  
\usepackage{courier}  
\usepackage{booktabs}
\usepackage{longtable}
\usepackage{subcaption}
\usepackage{amsmath}
\usepackage{verbatim}
\usepackage{array}
\usepackage{listings}
\usepackage{float}
\usepackage{tabularx}
\usepackage{xurl}
\usepackage{multirow}
\newif\ifcomment
\commenttrue
\ifcomment
\newcommand{\vr}[1]{{\textcolor{blue}{\textbf{VR: #1}}}}
\newcommand{\sz}[1]{{\textcolor{red}{\textbf{SZ: #1}}}}
\newcommand{\yz}[1]{{\textcolor{green}{\textbf{YZ: #1}}}}

\else
\newcommand{\vr}[1]{}
\newcommand{\sz}[1]{}
\newcommand{\yz}[1]{}
\fi

\newcommand\myparagraph[1]{%
  \noindent\textbf{#1}%
}

\usepackage[nameinlink,capitalize,noabbrev]{cleveref}

\usepackage{etoolbox}
\appto\appendix{%
  \crefalias{section}{appendix}%
  \crefalias{subsection}{appendix}%
  \crefalias{subsubsection}{appendix}%
}
\newcommand{\crefsub}[2]{\hyperref[#1]{\cref*{#1}#2}}

\usepackage{xcolor}

\usepackage[T1]{fontenc}

\usepackage[utf8]{inputenc}

\usepackage{microtype}

\usepackage{inconsolata}

\usepackage{graphicx}

\title{Expectation, Backlash, Recovery, and Excitement: How Model Releases Shape Reddit Perceptions of Conversational AI Systems}

\author{
  \textbf{Vahid Rahimzadeh}, \textbf{Yury Zhauniarovich}, \textbf{Savvas Zannettou} \\ 
  Delft University of Technology \\
  \texttt{\{v.rahimzadeh,y.zhauniarovich,s.zannettou\}@tudelft.nl} \\[0.7em]
  \href{https://vavre.github.io/p/upai}{\faRobot\ \texttt{vavre.github.io/p/upai}}
}

\begin{document}
\maketitle
\begingroup
\renewcommand{\thefootnote}{}
\footnotetext{This paper is appearing in the Proceedings of the 2026 Conference on Empirical Methods in Natural Language Processing (EMNLP 2026). Please cite the EMNLP version.}
\addtocounter{footnote}{-1}
\endgroup

\begin{abstract}

Conversational AI systems (CAISes) continuously change through model releases, feature updates, safety interventions, and access-policy shifts, yet user perceptions are often studied as static snapshots. We conduct a long-term, large-scale analysis of Reddit discussions to examine how users perceive CAIS model release interventions across providers. 
By combining sentiment classification and thematic concept analysis, we show that CAIS perceptions are dynamic and intervention-sensitive. Anthropic exhibits the clearest positive release profile through Claude Code and product-model fit, OpenAI shows backlash-and-recovery dynamics around GPT-5 and GPT-5.1, Grok-3 is shaped by provider identity and political discourse, and DeepSeek-R1 combines engineering praise with concerns about censorship, access, and reliability. These findings show that model releases are not merely technical updates, but user-facing interventions that reshape sentiment, expectations, and public discussion. 

\end{abstract}
\section{Introduction}
\begin{figure}[t]
  \centering
  \includegraphics[width=0.8\columnwidth]{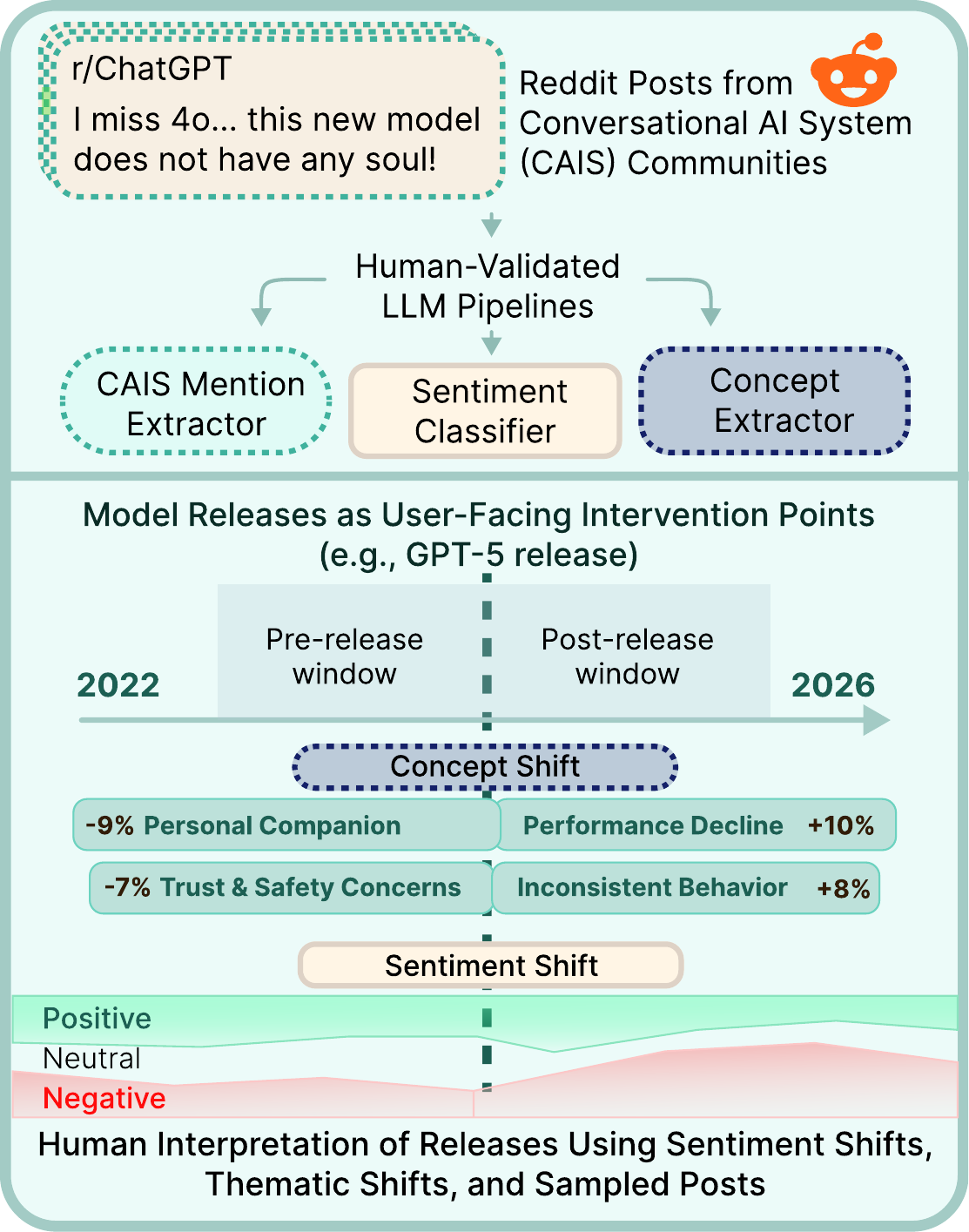}
\caption{Study overview of sentiment and thematic shifts around model releases.}  \label{fig:thumbnail}
\end{figure}

Conversational AI systems (CAISes) such as OpenAI's ChatGPT, Anthropic's Claude, Google's Gemini, and DeepSeek have rapidly become part of everyday life. Since the public release of ChatGPT in 2022, these systems are used daily by millions of users for tasks such as writing, information seeking, coding, emotional support, and other personal and professional activities~\cite{tamkin2024clio,tomlinson2025working,chatterji2025people,karnam2026bowling}, increasingly shaping how people access information, make decisions, and complete work. This widespread use makes it important to understand how people perceive CAISes. User perceptions shape whether people trust a system's outputs, share personal information with it, keep using a system, or abandon it after negative experiences. 

Prior work has examined user perceptions of CAISes mainly through user studies (e.g., surveys) and social media analysis~\cite{rauchfleisch2025culture_occupation_on_chatgpt,cheng2025metaphors}, identifying various factors such as perceived usefulness, trust, privacy concerns, and risks. Together, these previous studies provide important useful aggregate snapshots of how people use and perceive these systems; however, they are increasingly incomplete because CAISes are not stable. 
CAISes constantly change and evolve, for instance, through the introduction and deprecation of the underlying large language models, the introduction of features (e.g., memory feature~\cite{dash2026algorithmic}), safety updates, and changes to the rate limits. These interventions directly affect how users experience the CAIS, even when the product name remains the same. 
For instance, \citet{lai2026please} examined the ``\#Keep4o'' backlash and found that replacing the default model GPT-4o with GPT-5 in ChatGPT was experienced by users as a disruptive loss of functionality and a relationship with the system (i.e., users felt they had lost a friend). This case demonstrates that model changes can become socially meaningful events rather than merely technical updates. This example also points to a broader research gap. Existing work provides limited evidence on how users perceive and respond to different kinds of CAIS interventions across systems and providers. As a result, we know relatively little about whether reactions to system change are specific to exceptional backlash events (e.g., ``\#Keep4o'') or whether they reflect broader user experience.

Motivated by this research gap, we study user perceptions of CAIS interventions across multiple providers, focusing specifically on new model releases to understand how the introduction of newer models may affect user perceptions of CAISes. Specifically, we aim to provide answers to the following research questions.
\textbf{RQ1:} How does user sentiment about a CAIS change before and after a CAIS intervention on Reddit?
and \textbf{RQ2:} How do themes of discussion about a CAIS change before and after a CAIS intervention on Reddit?
We operationalize these questions using a release-centered design, summarized in \cref{fig:thumbnail}, that compares sentiment and thematic concept prevalence in symmetric windows before and after each model release.

To answer these questions, we collect a large-scale dataset of Reddit posts related to CAISes, covering 20 subreddits and 668K posts from October 2022 to December 2025. 
We develop human-validated LLM pipelines to identify model mentions and classify sentiment toward them. 
We also induce thematic concepts from the full corpus using a qualitative-research-inspired framework, producing interpretable concepts with explicit inclusion criteria. 
This allows us to use sentiment to identify where perceptions shift and thematic concepts to interpret what those shifts are about.

Our results show that user perceptions of CAISes are dynamic and intervention-sensitive. 
At the provider level, OpenAI and Google discussions become more critical over time, while Anthropic shows a distinctive positive trend. 
Around releases, Anthropic has the clearest positive intervention profile on average, whereas OpenAI shows more modest improvements and other providers are often closer to zero net change. 
Release-specific analysis reveals important differences: Claude-4 reflects strong product-model fit through Claude Code; GPT-4o discourse is shaped by future-oriented expectations and OpenAI's promises of improved capabilities; GPT-5 and GPT-5.1 illustrate backlash and recovery following a forced model change; Grok-3 shows how provider identity and political discussion shape CAIS perception; and DeepSeek-R1 shows how engineering praise can coexist with concerns about censorship, access, and reliability.

\section{Related Work}

\begin{figure*}[ht]
    \centering
    \includegraphics[width=\textwidth]{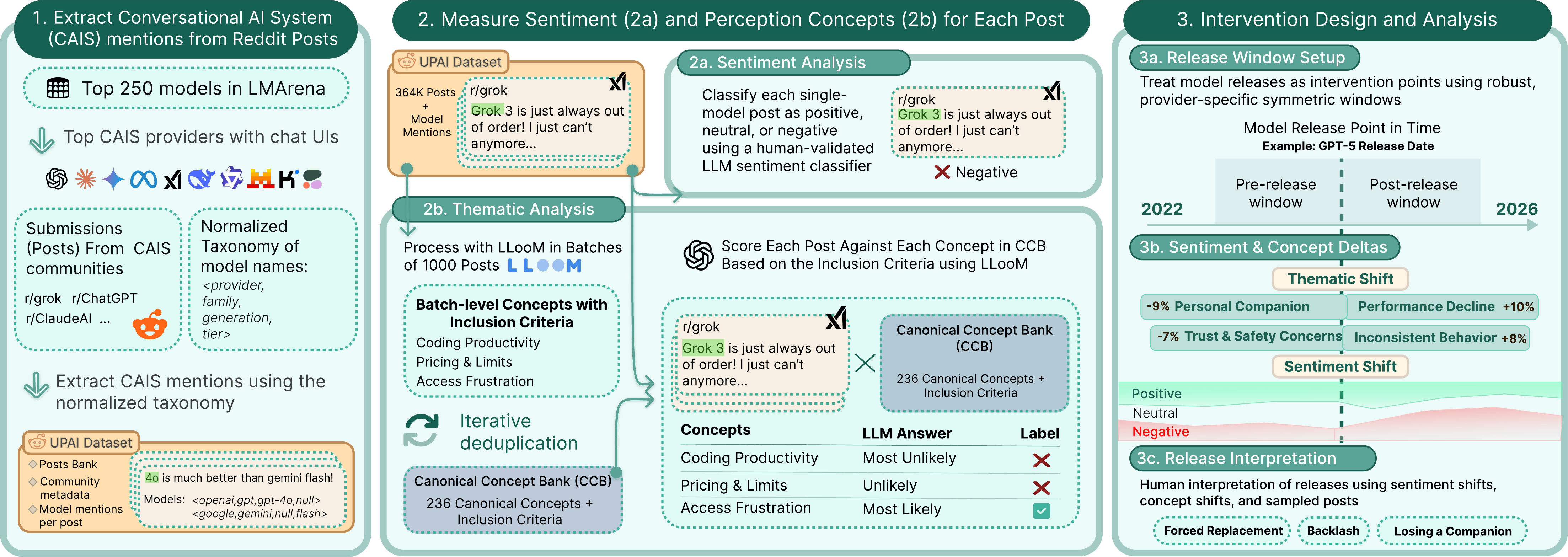}
\caption{Overview of our analysis pipeline. We ground model mentions in Reddit posts, measure post-level sentiment and perception concepts, and compare sentiment and concept prevalence in provider-specific windows around model releases.}
\label{fig:methodology}
\end{figure*}

Recent research focuses on understanding user perceptions of CAISes.
One branch examines how users broadly make sense of AI systems, showing that perceptions vary across explanations, external certification, metaphors, mental models, and everyday use~\cite{cheng2025metaphors,wang2025llm_mental_model,nylund2026llm_daily_life,cetinkaya2026between_explianibity_certificate_perception}. These studies show that users may understand AI systems as tools, collaborators, threats, or social actors, with consequences for trust, privacy concerns, reliance, and perceived value. A second body of works develops measures or taxonomies for particular perception constructs, including anthropomorphism~\cite{cheng2024anthroscore}, sentience-like interpretations in human--AI interaction~\cite{devrio2025anhtro_taxonomy}, trust in LLMs~\cite{de2025trust_llm_index}, and the attitudes towards LLMs generally and personal LLMs of use~\cite{liebherr2025general_specific_llm_attitude}. Together, these studies provide important conceptual and measurement foundations, but they do not fully capture how multiple perceptions evolve over time.
A growing body of large-scale empirical work studies public reactions to ChatGPT and generative AI across Reddit~\cite{xu2024reddit_chatgpt_23k_lda_vader,talafidaryani2024reddit_chatgpt_4k_lda_vader_sentistrenght_twopoint_data,qi2024reddit_genai_liwc_llm_sentiment_bertopi_33kcommnts}, Twitter/X~\cite{tounsi2023exploring_early_twitter_tool,koonchanok2024twitter_chatgpt_occupations,rauchfleisch2025culture_occupation_on_chatgpt}, Weibo/Bilibili~\cite{zhou2024chatgpt_chinese_knowledge_mapping}, cross-platform Twitter/Weibo comparisons~\cite{xu2026cross_weibo_twitter}, and app reviews~\cite{alabduljabbar2024genai_appstore_reviews}. These studies commonly combine LDA or BERTopic with lexicon-, transformer-, or LLM-based sentiment analysis, surfacing themes around technical use, prompting, education, labor, creativity, hallucination, regulation, and social impact. However, they report mixed sentiment patterns that vary by platform, time frame, community, and measurement method. Thus, while they demonstrate the value of large-scale public discourse, they often provide broad or aggregate trends. Related work on generative AI trust identifies longitudinal patterns in user concerns, but does not systematically compare release events~\cite{pessianzadeh2025generative_distrust}. 

Unlike works that capture aggregate snapshots, we treat model releases as recurring socio-technical events, inducing perception concepts from Reddit discussions, to systematically track how perception shifts across releases and providers. Closest to our framing, the \#Keep4o study, based on a qualitative analysis of 1.5K Twitter/X posts, shows that backlash to GPT-4o's replacement was driven by instrumental dependency and relational attachment~\cite{lai2026please}. While it richly explains one consequential event, here we study how such perception shifts can be systematically studied at scale.

\section{Dataset}
\label{sec:data_collection}
Our dataset relies on publicly available Reddit dumps (see \cref{app:reddit_dataset_details} for details). We collect all submissions made on Reddit between October 2022 and December 2025. 

\myparagraph{Subreddit Selection.}
We first identify AI providers and products using the LLM Arena text-to-text leaderboard as of November 28, 2025. We manually check the entire leaderboard and focus on retaining providers and products with public user-facing interfaces.
We identify two sets of providers: 
1)~\emph{Providers offering proprietary models:} OpenAI, Anthropic, Google, and xAI;
and 2)~\emph{Providers offering open-weight 
models:} like Meta, Mistral, Alibaba, and Moonshot AI. 
Also, we extract the names of popular CAISes from the leaderboard (e.g., ChatGPT, Gemini, etc.).
Then, using both the providers' and systems' names, we performed a manual search on the Reddit public search interface to identify relevant subreddits.
This process yields a set of 20 relevant subreddits (see \cref{tab:provider_subreddits}). 

\myparagraph{Filtering.}
We filter the collected Reddit dumps, retaining only submissions to the 20 subreddits, resulting in 849,603 submissions. For readability, we refer to Reddit submissions as posts throughout the paper.
We exclude all posts that were either removed or deleted (181K).
\cref{tab:per_subreddit} summarizes our corpus, which contains 668,063 posts.
\begin{table}[h]
\centering
\scalebox{0.85}{
\footnotesize
\setlength{\tabcolsep}{5pt}
\begin{tabular}{@{}lrr@{\hspace{10pt}}|@{\hspace{10pt}}lrr@{}}
\toprule
\textbf{Provider} & \textbf{\#Posts} & \textbf{\%} &
\textbf{Provider} & \textbf{\#Posts} & \textbf{\%} \\
\midrule
OpenAI    & 455,045 & 68.1 & DeepSeek & 9,982  & 1.5 \\
Meta      &  68,282 & 10.2 & Mistral  & 1,994  & 0.3 \\
Google    &  60,475 &  9.1 & Alibaba  &   976  & 0.1 \\
Anthropic &  43,271 &  6.5 & Moonshot &   223  & 0.0 \\
xAI       &  27,815 &  4.2 & \textbf{Total} & \textbf{668,063} & \textbf{100.0} \\
\bottomrule
\end{tabular}
}
\caption{Number of posts per provider in our data.}
\label{tab:per_subreddit}
\end{table}

\section{Methodology}
\label{sec:methodology}

Our methodology consists of three stages, summarized in \cref{fig:methodology}. First, we identify and normalize CAIS model mentions in Reddit posts using a taxonomy-guided LLM extractor. Second, we measure user perceptions at the post level through thematic concept analysis and sentiment classification. Finally, we analyze changes in sentiment and concept prevalence within provider-specific windows around model releases.

\subsection{Identifying CAIS Mentions}

Identifying mentions of specific providers and models is challenging because users may refer to the same model using different names (e.g., GPT-4, GPT4, ChatGPT4) or mention only the provider or product family. To address this, we construct a taxonomy of models from the providers considered in this study and use LLMs for mention identification.

\myparagraph{Taxonomy Creation.} We construct a four-level hierarchical taxonomy represented by a 4-tuple $<provider, family, generation, tier>$, where \emph{provider} is the releasing organization (e.g., Anthropic), \emph{family} is the product line (e.g., Claude), \emph{generation} is a specific release version (e.g., Claude-3-5), and \emph{tier} is a capability or size variant (e.g., Opus). 
To populate the taxonomy, we leverage the LLM Arena leaderboard, considering its top 250 entries.
We retain only entries associated with any of the 9 providers in our Reddit dataset (see \cref{tab:per_subreddit}) and derive a corresponding 4-tuple for each leaderboard entry. For example, the leaderboard entry \emph{llama-3.1-405B} is mapped to the 4-tuple $<meta, llama, 3.1, 405B>$. 
The final taxonomy comprises 189 entries spanning 9 providers, 15 model families, 50 generations, and 39 tier variants. More details on the process and final taxonomy are provided in \cref{app:model_taxonomy_details}.

\myparagraph{Identifying Model Mentions using LLMs.} 
To identify model mentions in Reddit posts, we employ an LLM that takes as input both our taxonomy and a Reddit post. The model is prompted to generate a JSON output indicating whether a post mentions a provider, family, generation, or tier. 
The LLM retains only mentions that can be mapped to entries in the taxonomy and is not allowed to infer models unless they are explicitly mentioned.
We provide the full prompts and more details on the parameters used during inference in \cref{app:loom_mention_hyperparams}.
Using this approach, we extract 364K posts that include 507K mentions (see \cref{tab:mention_distribution}).
Out of these, we exclude 2.1K mentions that do not follow our taxonomy; they either mention products outside of our taxonomy (e.g., Sora, Midjourney, etc.) or they provide some hallucinated output. In total, our final dataset consists of 505K model mentions from the Reddit dataset, spanned across 363K posts. Note that not all extracted mentions are associated with a full 4-tuple of provider, family, generation, and tier.
In particular, 41.5\% of the extracted mentions resolve only to a provider, 28.8\% resolve to a provider and a family, 21\% to a provider, family, and generation, while only 7.6\% of the mentions include a complete 4-tuple.
Also, 1.1\% of the extracted mentions include other combinations of missing entities, like including provider and generation, but not family and tier.
To validate the extractor, we manually annotated 397 randomly sampled posts, yielding 601 model mentions. The extractor exactly matches 83\% of raw textual model mentions and maps extracted mentions accurately to the taxonomy levels, with F1 scores of 1.00 for provider, 0.956 for family, 0.967 for generation, and 0.976 for tier. Full details on the annotation protocol, sample-size calculation, and validation results are in \cref{app:model_mention_validation}.

\begin{table}[t]
\centering
\scalebox{0.85}{
\footnotesize
\setlength{\tabcolsep}{5pt}
\begin{tabular}{@{}lrr@{}}
\toprule
\textbf{Provider} & \textbf{\#Mentions (\%)} & \textbf{\#Posts (\%)} \\
\midrule
OpenAI      & 316,410 (62.6) & 252,719 (69.5) \\
Google      &  61,793 (12.2) &  51,854 (14.3) \\
Anthropic   &  55,652 (11.0) &  44,020 (12.1) \\
xAI         &  20,075 (4.0)  &  18,418 (5.1)  \\
Meta        &  17,380 (3.4)  &  14,418 (4.0)  \\
DeepSeek    &  16,493 (3.3)  &  14,605 (4.0)  \\
Alibaba     &   9,518 (1.9)  &   7,209 (2.0)  \\
Mistral     &   6,942 (1.4)  &   5,801 (1.6)  \\
Moonshot AI &     987 (0.2)  &     884 (0.2)  \\
\midrule
\textbf{Total} & \textbf{505,250 (100.0)} & \textbf{363,546 (100.0)} \\
\bottomrule
\end{tabular}
}
\caption{Distribution of extracted mentions and posts.}
\label{tab:mention_distribution}
\end{table}

\subsection{Identifying User Perceptions}

To extract user perceptions from the posts associated with specific models, we adapt LLooM~\cite{lam2024concept}, an LLM-assisted framework inspired by qualitative research workflows. LLooM first extracts relevant excerpts from posts and summarizes them as bullet points. Then it clusters them into interpretable concepts with natural-language inclusion criteria, and finally scores each post against these concepts using a five-point Likert scale. This process enables the inductive discovery of auditable and human-interpretable perception concepts at scale, unlike fully qualitative methods~\cite{lai2026please}. At the same time, it avoids the limitations of topic modeling~\cite{xu2024reddit_chatgpt_23k_lda_vader} and taxonomy-based approaches~\cite{alabduljabbar2024genai_appstore_reviews}, which either require substantial post-hoc interpretation or rely on predefined categories that may overlook organically emerging concepts.

\myparagraph{Concept Generation and Deduplication.}
We apply LLooM with the seed prompt \emph{``How users perceive and use LLMs''} to the 363K posts containing at least one model mention. To fit within context-window constraints, we partition posts into 365 non-overlapping batches. Each batch is processed independently, yielding 5--30 raw concepts per batch (mean \(=13.6\)) and 4,979 raw concepts in total. 
\cref{app:loom_mention_hyperparams} provides more details about all LLooM parameters.
Because independent batches can produce duplicate or near-duplicate concepts, we implement an LLM-assisted deduplication procedure to construct a \textsc{Canonical Concept Bank} (CCB). E.g., \textit{Trust \& Reliability} appears across 96 batches as an exact-name duplicate, while concepts such as \textit{Creative Collaborator}, \textit{Creative Co-Creator}, and \textit{Creative Partner} use different names for closely related inclusion criteria. Our approach automatically merges exact-name duplicates, while remaining candidates are compared against existing CCB entries using both concept names and inclusion criteria. This reduces 4,979 raw concepts to 236 canonical concepts with inclusion criteria. Our prompt is provided in \cref{app:prompts} and CCB details are in \cref{app:concept_data_details}.

\myparagraph{Post Scoring using CCB.}
Next, we use the 236 unique concepts from the CCB to score all 363K posts in our dataset.
Note that the scoring function in LLooM compares each post to all generated concepts, making it both time- and resource-intensive.
To overcome this issue, we implement a batch-based scoring approach, where we score each post based on the concepts generated and extracted in that specific batch.
A concept appears in a post if LLooM's scoring function returns a verdict of ``Strongly Agree'' or ``Agree.'' 
Using this approach, we find that 20.6\% of the posts do not include any concepts from our CCB, 23.5\% include exactly one concept, and 48.9\% include between two and five concepts (see \cref{app:concept_data_details} for more details).

\subsection{Identifying Post Sentiments}
To measure post-level sentiment toward the mentioned model, we classify each single-mention post as positive, neutral, or negative using an LLM-based sentiment annotator. We validate this classifier on 150 human-annotated posts labeled by three annotators, obtaining a three-way nominal Krippendorff's $\alpha$ of 0.71. Against the majority label, the LLM achieves 0.82 weighted F1. We follow ~\cite{atreja2025prompt_design} for prompt design. We provide more details in \cref{app:sentiment_assignment_methodology}.

\subsection{Intervention Design}
\label{sec:intervention_design}

We treat model releases as provider-specific intervention points. Our analysis focuses on the six providers with more than 10K model mentions.
To reduce ambiguity in attributing perceptions to a model, we restrict release-window analyses to posts that mention only one model and contain a text body. 
For each release, we compare provider-specific Reddit discussions in symmetric windows before and after the release date. Because providers differ in release frequency, we use provider-specific window sizes (see \cref{app:intervention_design} for details). 
Concept-level changes are robust to this choice: recomputing them with windows down to $0.6\times$ the provider-specific size yields Pearson $r \geq .73$ with the reported values (see \cref{app:window_robustness} for details).

For each release, we compute two post-minus-pre deltas. First, we measure sentiment change as the difference in the share of positive, neutral, and negative posts before and after release. Second, we measure thematic change using CCB concept prevalence, defined as the share of posts in a window assigned to each concept. The concept delta is then the post-release prevalence minus the pre-release prevalence. To reduce noise from sparse concepts, we focus on concepts with at least 10 posts in the pre- or post-release window. 

\begin{figure*}[ht]
    \centering
    \includegraphics[width=\textwidth]{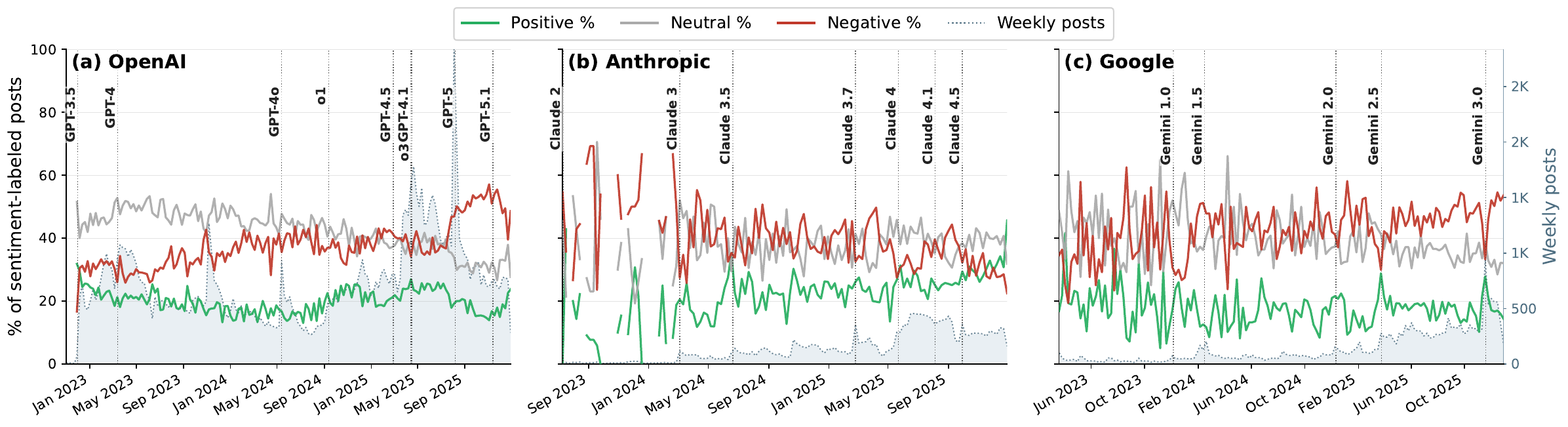}
    \caption{Weekly sentiment shares in Reddit posts mentioning a single CAIS for the three most-discussed providers. Asterisks mark FDR-corrected significance of each change ($^{}p<.05$, $^{}p<.01$, $^{}p<.001$).}
    \label{fig:provider-sentiment-timelines-3panel}
\end{figure*}

\section{Results}
\label{sec:results}

In this section, we investigate how Reddit users' perceptions of CAIS change around model releases, focusing on shifts in sentiment and evolving discussion themes.

\subsection{RQ1: Sentiment Changes}

\myparagraph{Overall Sentiment Trends.}
To characterize sentiment trends, we analyze positive, negative, and neutral posts separately, reporting both the change in the percentage of posts belonging to each class and the Theil--Sen trend estimates~\cite{hussain2019pymannkendall}. The latter is defined as the median slope across all pairs of points in a time series. 

Overall, Reddit discussions in our dataset skew negative: negative accounts for 38\% of posts, compared with 20\% positive and 42\% neutral, indicating that critical discussions of CAISes are nearly twice as common as favorable ones. 
Aggregated sentiment distributions across providers and model generations are reported in \cref{app:agreegated_sentiment}. 

\cref{fig:provider-sentiment-timelines-3panel} presents weekly sentiment trends for OpenAI, Anthropic, and Google, the three providers with the highest number of model mentions (see \cref{app:sentiment_trend_estimates} for all six providers).
OpenAI and Google both exhibit a shift from neutral toward more negative discussion. For OpenAI, neutral sentiment declines from roughly 45\% in late 2022 to 33\% by late 2025 (\(-0.117\) pp/week), while negative sentiment rises from 27\% to 46\% (\(+0.116\) pp/week), with little long-term change in positive sentiment (\(-0.008\) pp/week). This suggests that the overall shift is driven primarily by an erosion of neutral sentiment into more critical discussion. However, this pattern is diferent between GPT-4 and GPT-4o, where negative sentiment steadily increases while positive sentiment declines across the roughly 14-month period between the releases. Google follows a structurally similar but less steep trajectory: its negative sentiment rises more slowly than OpenAI's (\(+0.101\) vs. \(+0.116\) pp/week), and its neutral sentiment declines less steeply (\(-0.078\) vs. \(-0.117\) pp/week). However, Google also exhibits short-lived periods of positive convergence around the Gemini~2.5 and Gemini~3.0 releases. 

Anthropic is the main exception to this neutral-to-negative pattern. It is the only provider for which positive sentiment increases while negative sentiment declines significantly over time: from late 2023 to late 2025, positive sentiment rises from roughly 20\% to 35\%, while negative sentiment falls from 49\% to 27\%. Unlike OpenAI and Google, Anthropic therefore exhibits a long-term shift from negative toward positive sentiment.

\begin{figure}[t]
  \centering
  \includegraphics[width=\columnwidth]{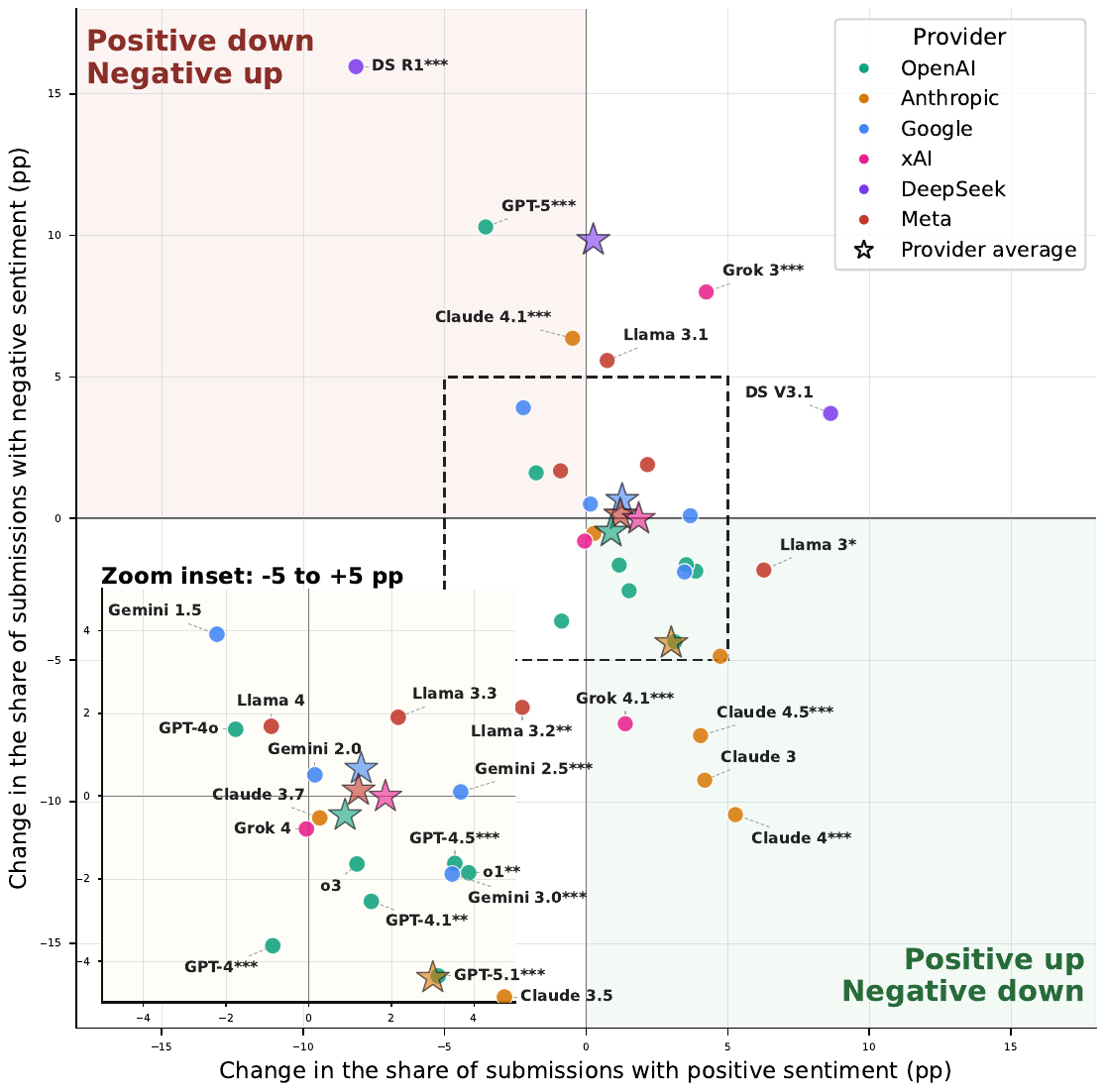}
\caption{Release-window changes in positive and negative sentiment for each model. Points show post-minus-pre sentiment changes; stars denote provider averages. 
The inset magnifies the dense central region (\(\pm 5\) pp). Asterisks on labels mark a statistically significant shift in the sentiment distribution ($^{}p<.05$, $^{}p<.01$, $^{}p<.001$).}
\label{fig:polarizing_model_releases}
\end{figure}

\myparagraph{Release-Specific Sentiment Shifts.}
\cref{fig:polarizing_model_releases} illustrates immediate sentiment shifts surrounding individual model releases (shown as dots), along with provider-level averages (shown as stars). Significance tests for all release-level sentiment shifts are provided in \cref{app:sentiment_significance}.

At the provider level, Anthropic stands out as the only provider whose releases are, on average, correlated with shifts that clearly fall within the favorable (bottom-right) quadrant.
On average, Anthropic releases are correlated with a \(+3.0/-4.4\) percentage-point (pp) change in positive/negative sentiment. OpenAI shows a smaller average change of \(+0.9/-0.5\) pp, while Google, Meta, and xAI remain closer to the origin. DeepSeek exhibits the least favorable average sentiment profile; however, this result is largely driven by the DeepSeek~R1 release, the most negative outlier in the dataset. These provider-level averages are descriptive summaries across releases; statistical significance is assessed at the individual-release level, with full results reported in \cref{app:sentiment_significance}. 

\begin{figure*}[ht]
    \centering
    \includegraphics[width=\textwidth]{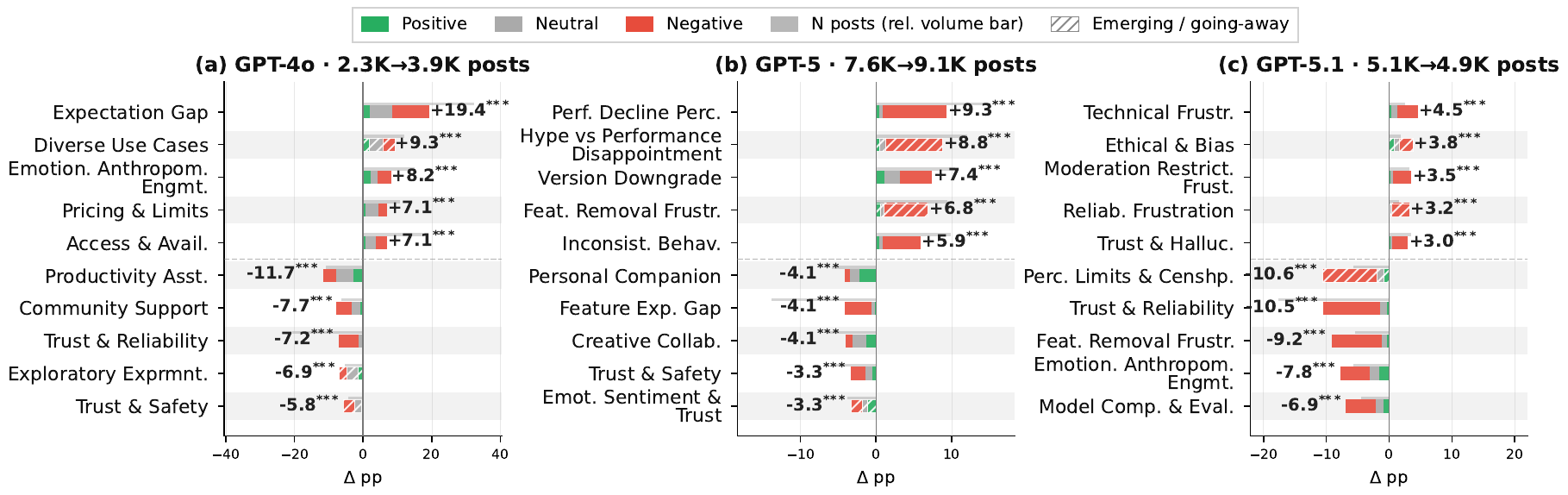}
    \caption{GPT-4o, GPT-5 and GPT-5.1 concept prevelance changes around releases. Asterisks mark FDR-corrected significance of each change ($^{}p<.05$, $^{}p<.01$, $^{}p<.001$).}
    \label{fig:openai-lineage-1x3}
\end{figure*}

Release-level results reveal substantial heterogeneity behind these provider averages. As noted earlier, DeepSeek~R1 is the most negative outlier in the dataset, with a \(-8.1\)/\(+16.0\) pp shift in positive/negative sentiment. Among OpenAI releases, GPT-5 release is correlated with the strongest negative shift (\(-3.5\)/\(+10.3\) pp), while within Anthropic, Claude~4.1 is the only release correlated with an unfavorable shift (\(-0.5\)/\(+6.4\) pp).
The strongest positive sentiment shifts are correlated with releases in the Claude family.
Claude~4 release correlates with the largest favorable shift (\(+5.3\)/\(-10.5\)), while Claude~3, Claude~3.5, and Claude~4.5 releases show similar patterns, increasing positive sentiment while reducing negative sentiment. By contrast, Grok~3 is a clear case of divided reception, as the release is correlated with increases in both positive and negative sentiment (\(+4.2\)/\(+8.0\)  pp).
Google's releases are more mixed: Gemini~2.5 is correlated with the largest positive shift among Google models (\(+3.7\)/\(+0.1\) pp).

\subsection{RQ2: Thematic Changes in Concepts}
We next examine which concepts become more or less salient around model releases. For each release, we compare concept prevalence before and after the release date and visualize the top five rising and falling concepts. To interpret these shifts, we manually inspect 20 randomly sampled posts per concept, which enables us to connect aggregate sentiment changes to the specific concerns, expectations, and use practices, changing around each release. We focus on releases with the most distinct sentiment profiles identified previously: GPT-5.1, GPT-5, GPT-4o, DeepSeek~R1, Claude~4, Grok~3, Claude~4.1, and Gemini~2.5. 
Due to space constraints, we discuss the first six releases in the main text and provide the analyses for the latter two in \cref{app:thematic_claude41_gemini25}. Paraphrased user quotes supporting the analyses are provided in \cref{app:thematic_quotes}, and significance tests for all concept changes in \cref{app:concept_significance}: of the 60 changes we discuss, 58 remain significant after FDR correction, the exceptions being two DeepSeek~R1 concepts affected by its small pre-release window.

\begin{figure*}[t]
    \centering
    \includegraphics[width=\textwidth]{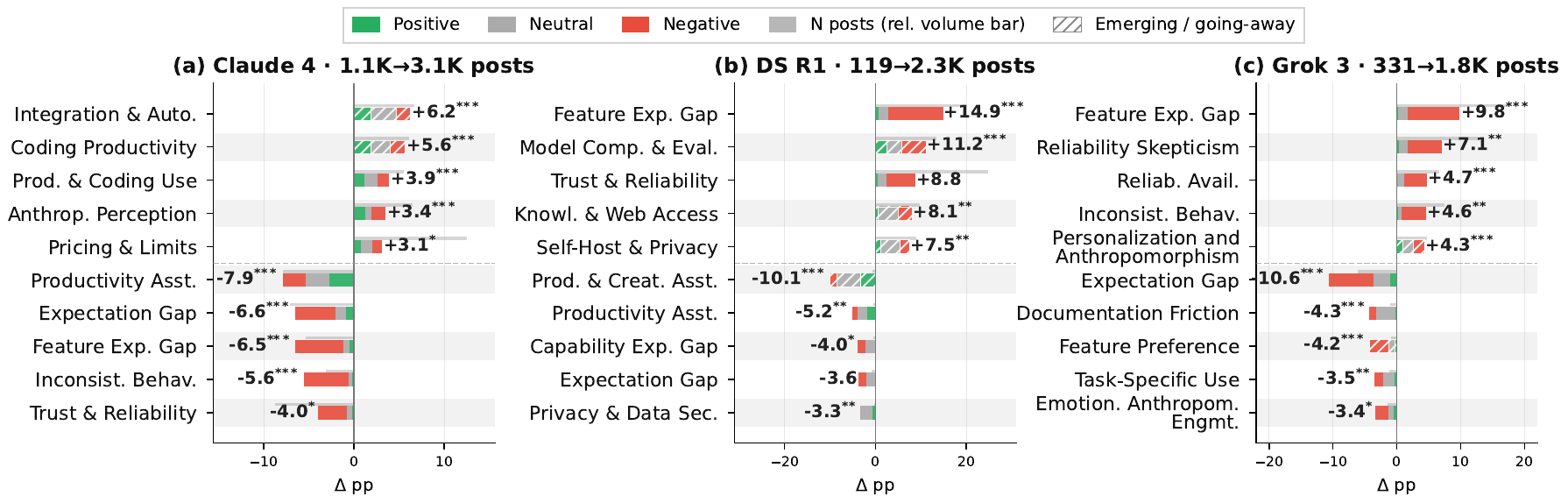}
    \caption{Claude-4, Deepseek-R1 and Grok-3 concept prevelance changes around releases.}
    \label{fig:claude4-dr1-grk3-1x3}
\end{figure*}

\myparagraph{GPT-4o: The Promised Future.}
GPT-4o is the clearest expectation-gap release in our analysis. As shown in \crefsub{fig:openai-lineage-1x3}{a}, \textit{Expectation Gap} increases by \(+19.4\) pp after release, the largest concept-level surge we observe in the entire data. This shift is not primarily about disappointment with the text model itself. Instead, users reacted to the distance between the highly visible ``omni'' demo and the capabilities actually available at launch. Posts repeatedly ask why real-time voice, image output, video/camera access, and interruption-based conversation were unavailable or limited despite being central to the launch presentation. In this sense, GPT-4o was received as a promised future that users could see but not yet fully access. This pattern is also reflected in the simultaneous rise of \textit{Access \& Availability} and \textit{Pricing \& Limits}: users were not only evaluating model quality, but trying to determine who could access the advertised capabilities, when, and under what subscription constraints.

GPT-4o also changes the emotional register of OpenAI discussion. \textit{Emotional Anthropomorphic Engagement} rises sharply, partly around voice interaction, ``her''-style references, and the Sky voice controversy \cite{altman2024her,openai2024skyvoice,reuters2024johansson}. However, this affective engagement is mixed rather than simply positive: users express excitement about more natural interaction while also questioning whether the released system matches the demo. At the same time, older GPT-4-era themes such as \textit{Productivity Assistant}, \textit{Exploratory Experimentation}, and \textit{Trust \& Reliability} decline. We interpret this as a displacement effect after roughly 14 months without a major OpenAI flagship release: mature discussions about everyday productivity recede, while discussion shifts toward multimodal promises, access constraints, and delivery gaps. Thus, GPT-4o generated intense attention, but its sentiment profile remained mixed because the central user question became not simply ``what can GPT-4o do?'', but ``why can I not use the capabilities that were shown?''

\myparagraph{GPT-5 and GPT-5.1: Emotional Backlash and Partial Recovery.}
GPT-5 and GPT-5.1 form a backlash-and-recovery sequence. GPT-5 is the strongest adverse OpenAI release in our analysis: as shown in \cref{fig:polarizing_model_releases},
positive sentiment decreases while negative sentiment rises sharply. Also, as shown in \crefsub{fig:openai-lineage-1x3}{b}, posts for all of the largest rising concepts after release are negatively discussed. 
This pattern aligns with the work from~\cite{lai2026please}, which identifies both relational attachment and instrumental dependency as central sources of user anger. Our concept shifts reveal both elements. On the relational side, backlash centers on \textit{Feature Removal Frustration} and a highly co-occurred concept \textit{Forced Upgrade Frustration}, where users frame the loss of GPT-4o as the removal of a familiar companion or emotionally meaningful interaction partner. On the instrumental side, the same release correlates with sharp increases in \textit{Performance Decline Perception}, \textit{Version Downgrade Expectation}, and \textit{Inconsistent Behavior Frustration}, where users report broken writing/coding workflows, memory continuity, and interaction styles. The emotional intensity of the response reflects both forms of disruption: GPT-5 was interpreted as a forced replacement for a system the users had already incorporated into their work and their emotional lives.
GPT-5.1 partially reverses this pattern, but its recovery is more about de-escalation than enthusiasm. Although several narrower complaints still rise after GPT-5.1 (\crefsub{fig:openai-lineage-1x3}{c}), their magnitudes are smaller than the large declines in GPT-5-era backlash concepts, including \textit{Feature Removal Frustration}, \textit{Trust \& Reliability}, \textit{Perceived Limitations \& Censorship}, and \textit{Emotional Anthropomorphic Engagement}. GPT-5.1 is therefore best understood as a partial recovery release: it improves sentiment by reducing the volume of GPT-5 backlash, shifting discussion away from platform-level revolt and toward narrower product complaints. 

\myparagraph{Claude-4: The Product-Model Fit.}
Claude~4 is the most favorable release-window shift in our analysis, correlating with an increase of positive sentiment by \(+5.3\) pp and  a reduction of negative sentiment by \(-10.5\) pp. As shown in \crefsub{fig:claude4-dr1-grk3-1x3}{a}, the post-window is dominated by coding-oriented concepts: \textit{Integration \& Automation}, \textit{Coding Productivity}, and \textit{Productivity \& Coding Use} are the largest rising concepts, with former two of them emerging from zero pre-window coverage. The main driver is Claude Code, which moved from a narrower research-preview with waitlist context around Claude~3.7 to public availability with Claude~4. In the sampled posts, Claude Code mentions rise sharply from 13/100 in the pre-window falling concepts to 64/100 in the post-window rising concepts, suggesting that users experienced Claude~4 not only as a model release but as a model--product bundle. The simultaneous decline of the broader \textit{Productivity Assistant} concept and rise of explicitly coding-focused concepts suggests that general productivity discussion was partly re-channeled into more specific coding and automation narratives once Claude Code became widely available. The release succeeds because model capability, product surface, and user workflow align around a clear use case: coding, automation, and agentic development support. 
The favorable reception coincides with an alignment between model capability, product surface, and user workflow around a clear use case: coding, automation, and agentic development support.
Before Claude~4, several declining concepts were dominated by negative discussion around capability degradation, inconsistent behavior, trust and reliability, and expectation gaps inherited from the Claude~3.7 and Max pricing plan period. After Claude~4, these failure-mode concepts recede, while coding-centered productivity narratives become more salient. However, the improvement is not complete: \textit{Pricing \& Limits} still rises after release, indicating that access and subscription grievances persist. 

\myparagraph{DeepSeek R1: Demand Shock.}
DeepSeek~R1 is the largest adverse release-window shift in our analysis, but its mechanism differs from the closed-model releases. As shown in \crefsub{fig:claude4-dr1-grk3-1x3}{b}, there is a discussion explosion rather than a replacement of an existing discourse: single-mention posts grow from 119 in the pre-window to 2,253 in the post-window, a roughly 19-fold increase. The dominant rising concepts point to demand overwhelming availability. \textit{Feature Expectation Gap}, \textit{Trust \& Reliability}, and \textit{Reliability Availability Frustration} all capture users encountering errors, downtime, and failed access shortly after release. In this sense, negative sentiment is driven less by disappointment with an established model and more by a newly expanded audience wanting to use R1 but repeatedly failing to access it. At the same time, DeepSeek~R1's reception is not uniformly negative. \textit{Model Comparison \& Evaluation} rises sharply and contains the strongest positive sentiment share among the rising concepts, reflecting admiration for R1's technical capability, price-performance profile, and challenge to OpenAI. Other emerging concepts reflect R1's distinctive positioning: \textit{Self-Hosting \& Privacy} captures neutral developer discussion around local deployment and distillation, while \textit{Bias \& Censorship Concerns} captures negative reactions to refusals or deletions around politically sensitive topics. Overall, R1 is best understood as a demand shock: a release that rapidly expanded public attention, but whose early reception was shaped by capacity failures, censorship discovery, and the gap between technical excitement and practical accessibility. 

\myparagraph{Grok-3: The Politicized Anthropomorphism.}
Grok~3 shows a distinctive divided-reception pattern: it is the only release in our analysis where both positive and negative sentiment increase meaningfully after release. As shown in \crefsub{fig:claude4-dr1-grk3-1x3}{c}, this mixed response is partly due to the expansion of the discussion itself, with the number of single-mention posts growing from 331 to 1,839. Posts associated with the most rising concepts are negative, including \textit{Feature Expectation Gap}, \textit{Reliability Skepticism}, \textit{Reliability Availability Frustration}, and \textit{Inconsistent Behavior Frustration}. These concepts capture a mixture of unmet expectations around Grok's ``maximally truth-seeking'' positioning, perceived quality decline, limits and access confusion, and operational failures. At the same time, Grok~3 also produces a notable rise in \textit{Personalization and Anthropomorphism}, an emerging concept in which users describe Grok through human-like agency, emotion, frustration, and persona-based interaction. What makes Grok~3 distinctive is that this anthropomorphic engagement is unusually politicized. Elon Musk appears repeatedly across rising concepts: as the source of the ``truth-seeking AI'' expectation, as the object of system-prompt and censorship debates, and as a political figure. As a result, Grok~3's reception cannot be reduced to model quality or product reliability alone. Its release becomes a site where users assess the model’s performance and attribute to it a political persona, institutional agenda, and human-like agency.
\section{Conclusion}
In this work, we examined how users perceive CAIS interventions through Reddit discussions surrounding major model releases across multiple providers. Our findings show that users perceive model releases not merely as technical upgrades, but as changes to familiar conversational systems that affect capability, reliability, trust, access, continuity, and emotional attachment. User reactions vary substantially across providers and release contexts, highlighting that static snapshots of perception are insufficient for continuously evolving systems. Taken together, our findings show that model releases are consequential user-facing events whose reception depends on both the model itself and the broader conditions surrounding its deployment.
Our results has several implications for providers and yield actionable recommendations. Specifically, capability announcements should be aligned with what users can actually access at launch, established models should remain available during transitions where possible, and infrastructure should be prepared for release-driven demand. Releases may also benefit from being paired with concrete workflows or use cases that make their practical value immediately visible, as observed around Claude 4 and Claude Code. Finally, because perceptions continue to evolve after release, providers should monitor user reactions over time and respond to emerging concerns rather than treating release reception as a one-time event.

\section*{Limitations}

Our study is subject to several potential limitations.

First, our analysis is limited to text-based Reddit posts. We do not analyze posts that contain only multimedia content, such as image-only or video-only posts, which may also contain relevant information about user perceptions of CAISes. This limitation stems from the dataset used in this study, which includes only textual information. As a result, our findings capture only the aspects of user discussion that are expressed in text.

Second, Reddit users do not necessarily represent the broader population of conversational AI users. Reddit communities discussing CAISes are likely to include more technically engaged and early-adopting users, whose expectations, concerns, and usage practices may differ from those of general users. Therefore, our findings should be interpreted as reflecting perceptions within these online communities rather than as a general measure of public opinion.

Third, some Reddit posts may themselves be AI-generated or AI-assisted, which we cannot reliably identify from the available data. Consequently, our analysis reflects the discourse present on Reddit and may not exclusively capture human-authored perceptions.

Fourth, our use of the LLooM framework required splitting the dataset into batches, which were analyzed independently and then post-processed to aggregate the resulting concepts. While this design allowed us to scale the method to a large corpus, it may have introduced additional redundancy into the final concept list. Despite our efforts to mitigate common errors during the preliminary experimental design, the resulting CCB may contain more concepts than would have emerged if all posts had been analyzed jointly in a single batch.

Fifth, our analysis of sentiment and concept shifts relies on symmetric time windows around model release dates, with window sizes defined by the statistics of release gaps for each provider. While this choice provides a consistent comparison framework, it may smooth out short-lived reactions that occur immediately after a release. For example, the case of Gemini~2.5 suggests that some sentiment shifts persist for only several weeks, whereas the average analysis window for Google spans 104 days. Nevertheless, major and sustained changes should remain visible within the longer observation period and therefore be captured by our approach. This is supported by the fact that our method captures several already-known major events.

Finally, our pipeline relies on LLM-based methods to identify model mentions, classify sentiment, and associate posts with perception concepts. These methods enable scalable analysis, and we evaluate them using annotated validation sets created by multiple annotators; however, they are not perfectly accurate and may introduce errors into the results. Consequently, some model mentions, sentiment labels, or concept assignments may be missed or incorrectly classified, which should be considered when interpreting the findings.

\section*{Ethical Considerations}
We emphasize that our work relies on analyzing data that are exclusively shared in a publicly available platform (i.e., Reddit).
When conducting our analysis, we follow standard ethical guidelines such as undertaking analysis in an aggregate manner, not trying to deanonymize users, or track them across websites.

Despite the public nature of the dataset, we ensured that the user's privacy is protected. 
Specifically, we opted for using exclusively open-source LLMs for our analysis, which means that no user data was disclosed to any provider offering LLMs.
Also, when reporting user quotes to substantiate our claims and findings, we paid particular attention to paraphrasing the quotes so that the specific quotes can not be linked to a specific Reddit user.
Overall, we do not anticipate any potential misuse of the presented results or analysis.

\bibliography{custom}

\appendix

\section{Dataset Details}
\label{app:dataset_details}

This appendix section provides additional details about the construction and characterization of datasets collected and generated in different steps of our work. 
\cref{app:reddit_dataset_details} describes the external data sources, subreddit selection process, and provider-level subreddit coverage for the initial Reddit dataset. 
\cref{app:model_mention_details} reports additional statistics on provider mentions and co-mentions. 
\cref{app:concept_data_details} summarizes concept-assignment statistics and provides the concept inclusion criteria used in the intervention analysis.

\subsection{Reddit Dataset}
\label{app:reddit_dataset_details}
This subsection describes the data sources and subreddit selection procedure used to construct the Reddit dataset. 
\cref{tab:data_sources} lists the external sources used in the collection process, including the Reddit dumps and the LLM Arena text-to-text leaderboard. 
\cref{tab:provider_subreddits} reports the subreddits identified for each conversational AI provider during subreddit selection. 
Three providers identified during the leaderboard review, Baidu, Zhipu AI, and Meituan, had no dedicated Reddit presence at the time of curation and were therefore excluded. 
Overall, a substantial share of activity in our dataset comes from OpenAI-related communities, with 68.1\% of submissions posted in \texttt{r/ChatGPT} and \texttt{r/OpenAI}.

\begin{table}[!htb]
\centering
\scalebox{0.85}{
\footnotesize
\setlength{\tabcolsep}{5pt}
\begin{tabular}{@{}l >{\raggedright\arraybackslash}p{0.9\columnwidth} @{}}
\toprule
\textbf{Provider} & \textbf{Subreddits} \\
\midrule
OpenAI & \texttt{r/ChatGPT}, \texttt{r/OpenAI} \\
Anthropic & \texttt{r/ClaudeAI}, \texttt{r/Anthropic}, 
\texttt{r/AnthropicAi} \\
Google & \texttt{r/GeminiAI}, \texttt{r/GoogleGeminiAI}, \texttt{r/GoogleGemini}, \texttt{r/Bard}, \texttt{r/GoogleBard} \\
DeepSeek & \texttt{r/DeepSeek} \\
xAI & \texttt{r/grok}, \texttt{r/GrokAI}, \texttt{r/xai}, \texttt{r/xAI\_community} \\
Meta & \texttt{r/MetaAI}, \texttt{r/LocalLLaMA} \\
Mistral & \texttt{r/MistralAI} \\
Alibaba & \texttt{r/Qwen\_AI} \\ 
Moonshot & \texttt{r/kimi} \\
\bottomrule
\end{tabular}
}
\caption{Identified subreddits per conversational AI provider.}
\label{tab:provider_subreddits}
\end{table}

\begin{table}[!htb]
\centering
\scalebox{0.85}{
\footnotesize
\setlength{\tabcolsep}{5pt}
\begin{tabular}{@{}p{1.5cm}p{4cm}p{2.6cm}@{}}
\toprule
\textbf{Source} & \textbf{URL} & \textbf{Use} \\
\midrule
Reddit dumps &
\url{https://academictorrents.com/details/3d426c47c767d40f82c7ef0f47c3acacedd2bf44} &
Collection of Reddit submissions \\

LLM Arena leaderboard &
\url{https://lmarena.ai/leaderboard/text} &
Identification of providers and CAISes \\
\bottomrule
\end{tabular}
}
\caption{External data sources used in the dataset construction.}
\label{tab:data_sources}
\end{table}

\subsection{CAIS Mentions}
\label{app:model_mention_details}
\begin{figure}[!htb]
  \centering
  \includegraphics[width=\columnwidth]{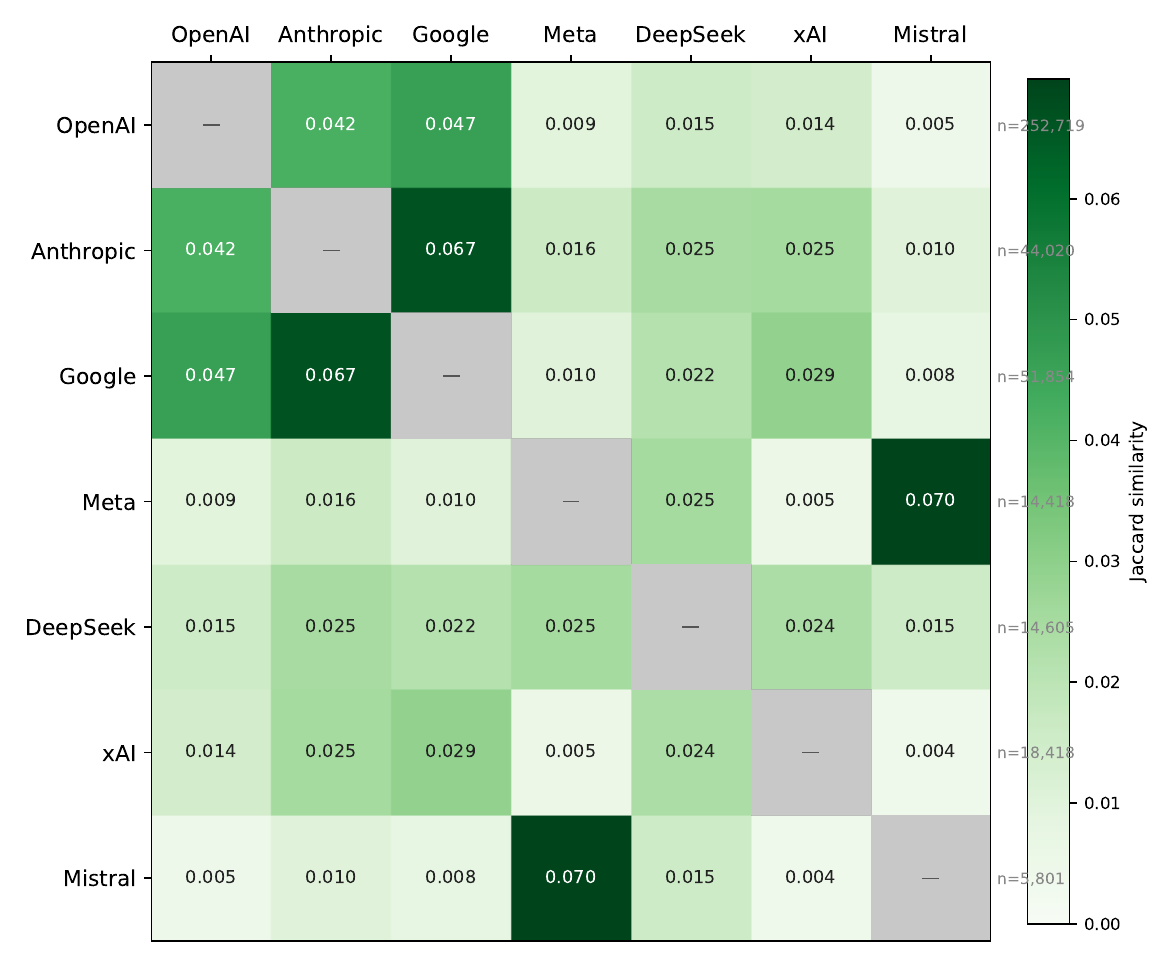}
  \caption{Pairwise Jaccard similarity between providers computed over
           post-level co-mentions ($|A \cap B| / |A \cup B|$).
           Row labels show each provider's total post count~$n$.
           Higher values indicate a larger share of posts mentioning
           both providers relative to either alone.}
  \label{fig:provider_comention_jaccard}
\end{figure}

\begin{figure}[!htb]
  \centering
  \includegraphics[width=\columnwidth]{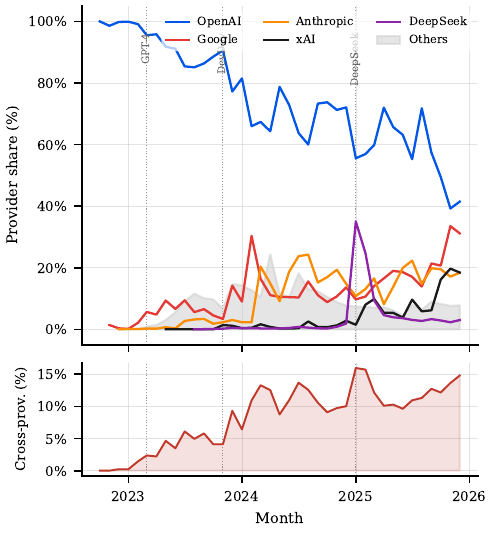}
  \caption{Monthly provider attention share (\%) across the top-5 providers
           (top panel) and cross-provider co-mention rate (bottom panel).
           }
  \label{fig:provider_share_hhi}
\end{figure}

\begin{figure}[!htb]
  \centering
  \includegraphics[width=\columnwidth]{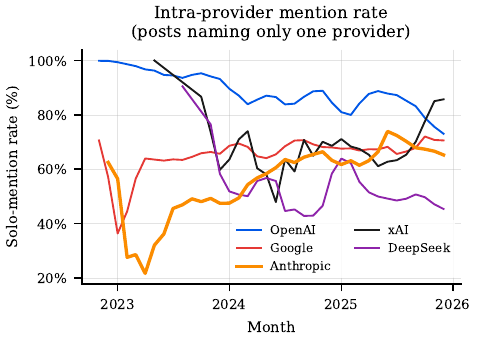}
  \caption{Monthly solo-mention rate per provider: the fraction of a
           provider's posts that name only that provider, smoothed with a
           3-month rolling mean. Declining rates indicate increasing
           comparative discussion.}
  \label{fig:intra_provider_trend}
\end{figure}

This subsection provides additional descriptive statistics for provider mentions and co-mentions in the dataset. 
\cref{fig:provider_comention_jaccard} shows pairwise Jaccard similarity between providers based on post-level co-mention overlap. 
Anthropic--Google is the strongest closed-provider pair (0.067), reflecting frequent comparative discussion of frontier closed models, while Meta--Mistral stands out as the highest cross-tier pair (0.070), driven by open-source community discussions of both providers. 
\cref{fig:provider_share_hhi} shows monthly provider attention share and cross-provider co-mention rate over time, and \cref{fig:intra_provider_trend} shows the intra-provider discussion rate per provider across the observation window. Intra-provider means conversations that incorporate only one provider, whether talking about one model or different models of the same provider.

\subsection{User Perceptions and Concepts}
\label{app:concept_data_details}

\begin{figure}[!htb]
  \centering
  \includegraphics[width=\columnwidth]{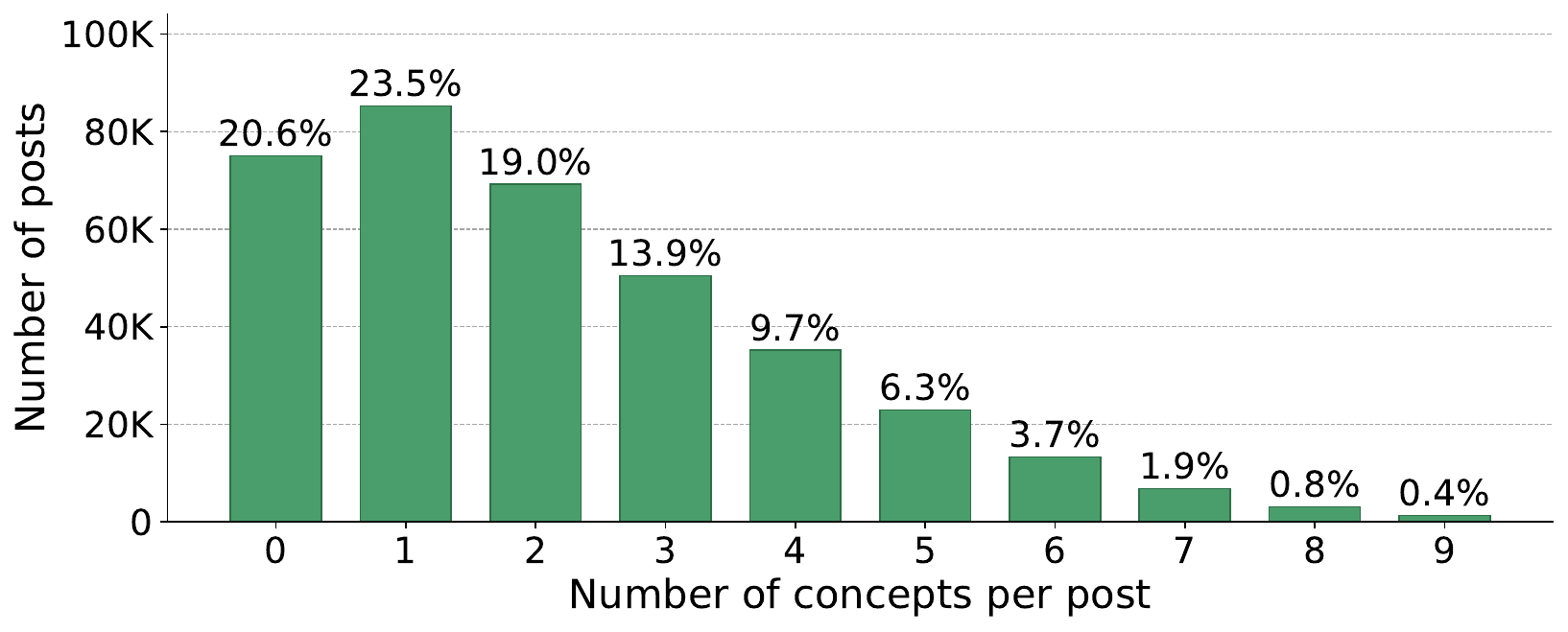}
  \caption{Distribution of number of concepts per post.}
  \label{fig:concepts_per_post}
\end{figure}

\begin{figure*}[!htb]
  \centering
  \includegraphics[width=\textwidth]{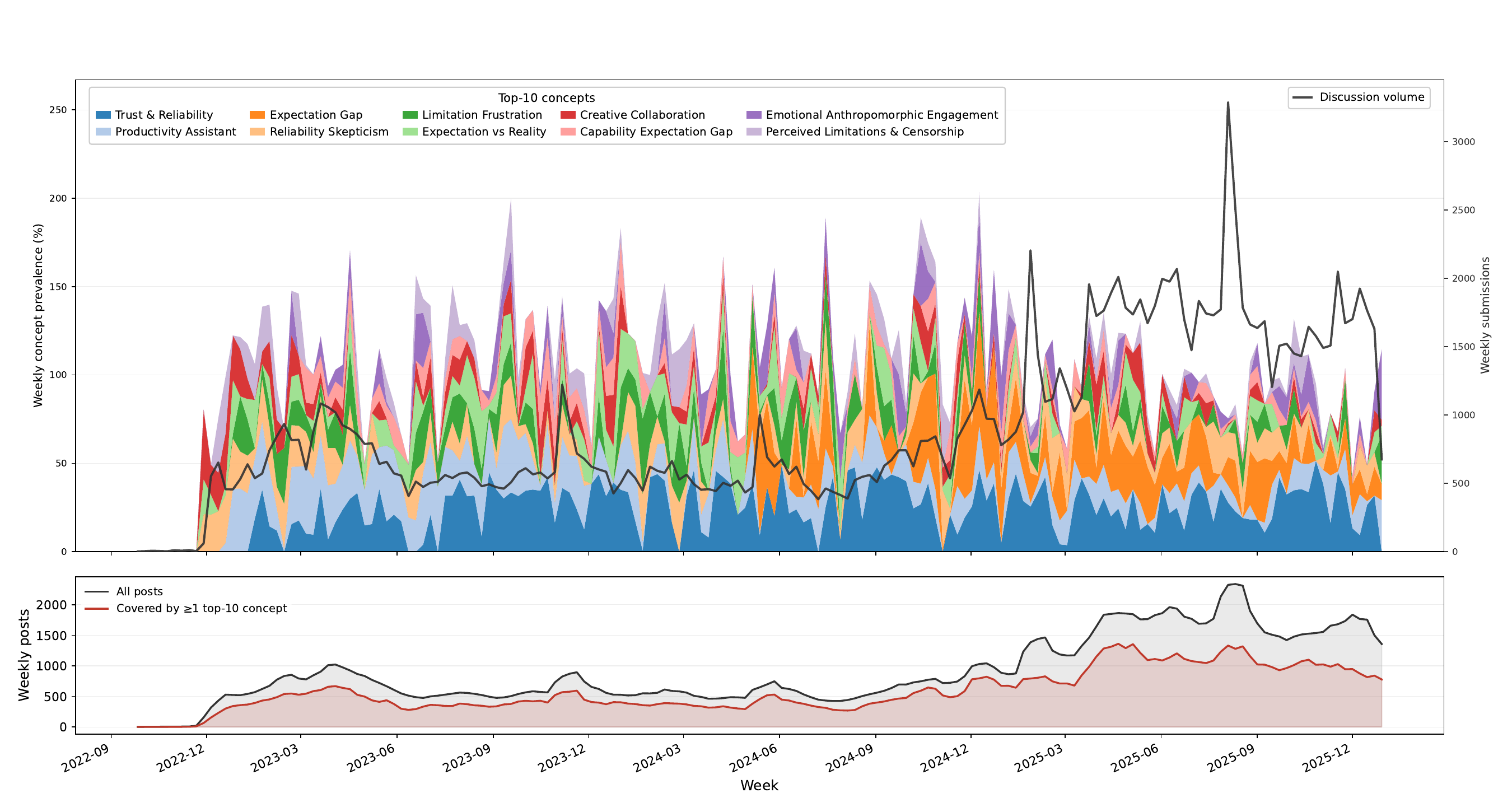}
  \caption{Weekly prevalence of the top-10 concepts (stacked area, left axis)
           and total discussion volume (line, right axis) across the full
           observation window, using the single-mention / no-link filtered
           corpus ($n=185{,}181$ posts).
           The bottom panel shows total weekly posts (grey) against posts
           covered by at least one top-10 concept (red).}
  \label{fig:concept_area_top10}
\end{figure*}

\begin{figure*}[!htb]
    \centering
    \includegraphics[
        width=\textwidth,
        height=0.92\textheight,
        keepaspectratio
    ]{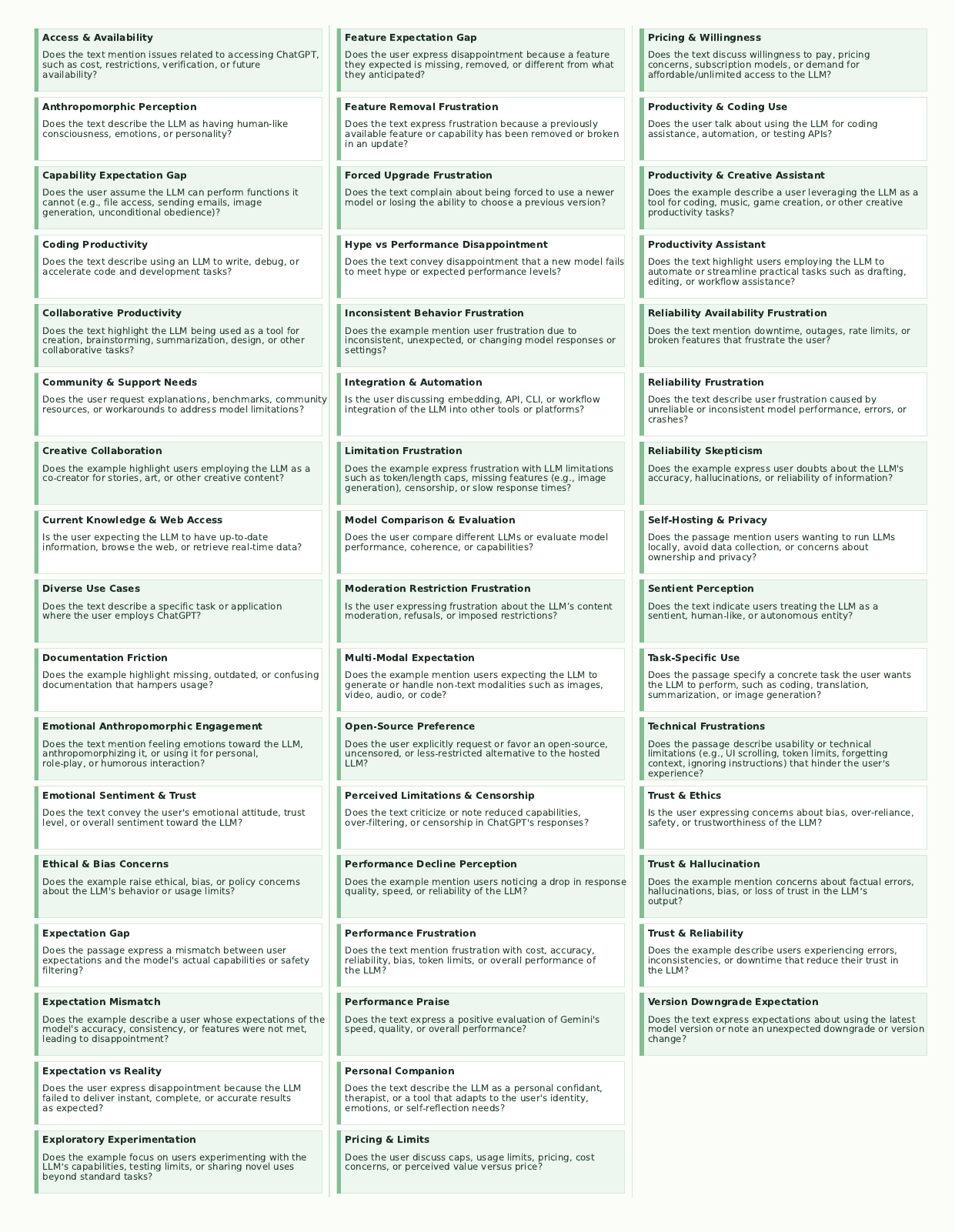}
    \caption{Concept names and inclusion criteria used for the selected top-5 intervention plots.}
    \label{fig:selected_intervention_concept_criteria}
\end{figure*}

This subsection provides additional details about concept assignments and the Canonical Concept Bank (CCB). 
\cref{fig:concepts_per_post} shows the distribution of the number of concepts assigned per post, suggesting a right-skewed distribution in number of concepts assigned per post. 
\cref{fig:concept_area_top10} shows the longitudinal prevalence of the top-10 concepts ranked by weighted weekly visibility, stacked as area shares of weekly post volume, together with the number of posts covered by at least one top-10 concept. An interesting observation is the emergence of "Expectation Gap" around GPT-4o release and staying in trend for the rest of the time period.
\cref{fig:selected_intervention_concept_criteria} reports the subset of CCB concept names and inclusion criteria used for the selected intervention analyses.

\section{Methodology Details}
\label{app:methodology_details}

This appendix provides additional implementation and validation details for the methodological components used in the paper. 
\cref{app:lloom_mention_extraction_metholodgy} describes the experimental setup, model-mention extraction pipeline, taxonomy construction, and prompts used for mention extraction and concept induction. 
\cref{app:sentiment_assignment_methodology} describes the sentiment annotation protocol, validation set, classifier performance, and sentiment prompt. 
\cref{app:intervention_design} describes the release-window design and the computation of sentiment and concept deltas used in the intervention analysis.

\subsection{Identifying CAIS Mentions and User Perceptions}
\label{app:lloom_mention_extraction_metholodgy}

This subsection provides implementation details for two components of our pipeline: model mention extraction and LLooM-based concept induction. 
Across these experiments, we use \texttt{openai/gpt-oss-120b} served locally on 2$\times$H200 GPUs.

\subsubsection{Taxonomy Creation}

\label{app:model_taxonomy_details}

This subsection describes how raw model names were normalized into the shared model taxonomy used by the mention extractor.
We constructed the model taxonomy through an iterative refinement process involving three researchers. Starting from the retained LLM Arena leaderboard entries, each model name was mapped to a four-level tuple (<provider, family, generation, tier>). Because providers use different naming conventions, the researchers reviewed and refined the mappings to ensure consistency across providers. For example, model names that encode size, capability, or deployment variants differently across providers were normalized into the shared \emph{tier} field, while product-line names were normalized into the \emph{family} field. This process produced a final taxonomy of 189 entries spanning 9 providers, 15 model families, 50 generations, and 39 tier variants.
\cref{tab:model_taxonomy_examples} shows examples of how model names are normalized into this shared taxonomy. These examples illustrate how provider-specific naming conventions are mapped into a common (<provider, family, generation, tier>) structure. The full taxonomy is available in our anonymous repository.\footnote{\url{https://anonymous.4open.science/r/user_perception_ai_anon-0FBB}}

\begin{table}[t]
\centering
\scalebox{0.8}{
\footnotesize
\setlength{\tabcolsep}{3pt}
\begin{tabular}{ll}
\toprule
Model & Normalized tuple \\
\midrule
Qwen3 Max Preview & alibaba / qwen / qwen-3 / max \\
Qwen3 235B Instruct & alibaba / qwen / qwen-3 / 235b \\
Qwen2.5 Max & alibaba / qwen / qwen-2-5 / max \\
\midrule
Claude Sonnet 4.5 & anthropic / claude / claude-4-5 / sonnet \\
Claude Opus 4.1 & anthropic / claude / claude-4-1 / opus \\
Claude 3 Haiku & anthropic / claude / claude-3 / haiku \\
\midrule
DeepSeek R1 & deepseek / deepseek-r1 / deepseek-r1 / base \\
\midrule
Gemini 3 Pro & google / gemini / gemini-3-0 / pro \\
Gemini 2.5 Pro & google / gemini / gemini-2-5 / pro \\
Gemini 2.5 Flash & google / gemini / gemini-2-5 / flash \\
\midrule
Llama 2 7B & meta / llama / llama-2 / 7b \\
Llama 3.2 3B & meta / llama / llama-3-2 / 3b \\
\midrule
Kimi K2 Thinking Turbo & moonshot-ai / kimi / kimi-k2 / turbo \\
\midrule
GPT-4o Mini & openai / gpt / gpt-4o / mini \\
GPT-5 Chat & openai / gpt / gpt-5 / base \\
GPT-4.1 Mini & openai / gpt / gpt-4-1 / mini \\
\midrule
Grok 4.1 & xai / grok / grok-4-1 / base \\
\bottomrule
\end{tabular}
}
\caption{Examples of model taxonomy normalization.}
\label{tab:model_taxonomy_examples}
\end{table}

\subsubsection{Hyperparameters and Experimental Setup}
\label{app:loom_mention_hyperparams}
\cref{tab:app_all_settings} summarizes the inference and runtime settings for the main pipeline components. 
Unless otherwise specified, LLooM hyperparameters follow the default settings of the \texttt{text-lloom} library.
\begin{table}[!htb]
\centering
\scalebox{0.85}{
\footnotesize
\begin{tabularx}{\columnwidth}{@{}lX@{}}
\toprule
\multicolumn{2}{@{}l}{\textbf{A.\ Model Mention Extraction}} \\
\midrule
Temperature         & 0.0 \\
Max tokens          & 3,096 \\
Timeout             & 180\,s \\
Max retries         & 5 \\
Concurrent workers  & 128 \\
\midrule
\multicolumn{2}{@{}l}{\textbf{B.\ Concept Scoring}} \\
\midrule
Temperature         & 0.0 \\
Relevance threshold & agree/strongly agree \\
Post--concept pairs & 4,605,080 \\
Assignment type     & Non-exclusive (multi-label) \\
\midrule
\multicolumn{2}{@{}l}{\textbf{C.\ LLooM Concept Generation}} \\
\midrule
Temperature         & 0.0 \\
Max tokens          & 6,000 \\
Embedding model     & \texttt{codefuse-ai/F2LLM-4B} \\
Embedding device    & H200 \\
Parallel workers    & 10 \\
Total batches       & 365 \\
Failed batches      & 0 \\
\midrule
\multicolumn{2}{@{}l}{\textbf{D.\ Deduplication (LLM variants)}} \\
\midrule
Temperature         & 0.0 \\
Max tokens          & 10,000 \\
\bottomrule
\end{tabularx}
}
\caption{Hyperparameters and experimental setups for concept generation and model mention extraction.}
\label{tab:app_all_settings}
\end{table}


\subsubsection{Model Mention Extraction Validation}
\label{app:model_mention_validation}

This subsection validates the model-mention extraction pipeline against a manually annotated sample. To validate our mention extractor, we randomly sampled 397 posts from the full dataset of 668K posts. This sample size is statistically representative at a 95\% confidence level with a 5\% margin of error. The manual annotation process followed the same instructions as the mention extractor: given the taxonomy and a Reddit post, the annotator identified all model mentions, recorded the raw textual mention, and mapped each mention to the corresponding 4-tuple.

The annotation was performed independently by one author without access to the extractor outputs. The resulting gold set contains 601 raw model mentions and corresponding taxonomy mappings. Comparing the extractor output with this gold set, we find that 501 raw model mentions (83\%) are exact matches. We then evaluate mapping quality at each taxonomy level. As shown in \cref{tab:evaluation_results}, the extractor achieves weighted F1 scores of 1.00 for provider, 0.956 for family, 0.967 for generation, and 0.976 for tier. These results indicate that while the extractor may miss some mentions, the mentions it extracts are mapped accurately to the taxonomy.

\begin{table}[!htb]
\centering
\scalebox{0.85}{
\footnotesize
\setlength{\tabcolsep}{5pt}
\begin{tabular}{@{}lrrrr@{}}
\toprule

\textbf{Taxonomy Level} & \textbf{Acc.} & \textbf{Prec.} & \textbf{Rec.} & \textbf{Weighted\,F1} \\

\midrule

Provider   & 1.000 & 1.000 & 1.000 & 1.000 \\

Family     & 0.958 & 0.901 & 0.956 & 0.956 \\

Generation & 0.960 & 0.763 & 0.967 & 0.967 \\

Tier       & 0.980 & 0.773 & 0.976 & 0.976 \\

\bottomrule

\end{tabular}
}
\caption{Mention mapping to taxonomy 4-tuples performance. We use Weighted F1 score due to class imbalance across taxonomy levels.}
\label{tab:evaluation_results}
\end{table}

\subsubsection{Prompts}

\label{app:prompts}
This subsection reports the prompts used for model mention extraction, LLooM concept induction, concept scoring, and concept deduplication.
The prompt in \cref{fig:prompt_extraction} is used verbatim for model mention extraction; \texttt{\_\_taxonomy\_HIERARCHY\_\_} is substituted with the finalized model taxonomy at runtime and \texttt{\{reddit\_submission\_text\}} with the concatenated title and selftext.

\begin{figure*}[!htb]
\begin{lstlisting}[basicstyle=\scriptsize\ttfamily, breaklines=true, frame=single]
You are a Literal Entity Extractor for a longitudinal study on AI models.
Your goal is to extract model mentions and normalize each mention to:
provider -> family -> generation -> tier

### 4-LEVEL SCHEMA
1. `provider`: Organization slug (example: `openai`, `meta`)
2. `family`: Family slug within provider (example: `gpt`, `o`, `claude`, `llama`)
3. `generation`: Generation slug within family
4. `tier`: Variant/size/capability slug within generation

### VALID HIERARCHY (STRICT MATCHING)
Use ONLY the valid hierarchy below.
__TAXONOMY_HIERARCHY__

### EXTRACTION RULES
1. Extract only explicit model mentions from the text.
2. Return only ONE object per unique model mention string in a post (case-insensitive, trimmed).
3. If the same mention appears multiple times, keep only one instance (prefer the first span in text).
4. Do NOT output overlapping substring duplicates. If `ChatGPT Plus` is the mention, do not also emit `ChatGPT` from that same span.
5. Do NOT infer unsupported details. If a level is not explicit, return null for that level.
6. Keep parent-child consistency with the hierarchy.
7. `raw_mention` must be the exact text span from input.
8. Prefer the minimal core entity span. Exclude non-entity trailing descriptors.
9. Do NOT emit pure provider/company references (e.g. `OpenAI`, `Google`) unless directly used as a model name.
10. Accept common misspellings when context clearly points to a known model.
11. Output JSON only, no explanation.

### INPUT TEXT
"{reddit_submission_text}"

### OUTPUT FORMAT (JSON)
{"mentions": [{"raw_mention": "substring from input", "provider": "slug or null",
               "family": "slug or null", "generation": "slug or null", "tier": "slug or null"}]}
\end{lstlisting}
\caption{Model mention extraction prompt. Template fields are substituted at runtime.}
\label{fig:prompt_extraction}
\end{figure*}

LLooM~\cite{lam2024concept} uses four prompt templates from the \texttt{text-lloom} library, reproduced below. Template fields in braces are filled at runtime by the framework.

\begin{figure*}[!htb]
\begin{lstlisting}[basicstyle=\scriptsize\ttfamily, breaklines=true, frame=single,
    title={Stage 1a -- Distill: Quote Extraction}]
I have the following TEXT EXAMPLE:
{ex}

Please extract {n_quotes} QUOTES exactly copied from this EXAMPLE that are {seeding_phrase}.
Please respond ONLY with a valid JSON:
{"relevant_quotes": ["<QUOTE_1>", "<QUOTE_2>", ...]}
\end{lstlisting}
\begin{lstlisting}[basicstyle=\scriptsize\ttfamily, breaklines=true, frame=single,
    title={Stage 1b -- Distill: Summarize}]
I have the following TEXT EXAMPLE:
{ex}

Please summarize the main point of this EXAMPLE {seeding_phrase} into {n_bullets} bullet
points, where each bullet point is a {n_words} word phrase.
Please respond ONLY with a valid JSON:
{"bullets": ["<BULLET_1>", "<BULLET_2>", ...]}
\end{lstlisting}
\begin{lstlisting}[basicstyle=\scriptsize\ttfamily, breaklines=true, frame=single,
    title={Stage 2 -- Synthesize: Concept Induction (produces name + inclusion criterion)}]
I have this set of bullet point summaries of text examples:
{examples}

Please write a summary of {n_concepts_phrase} for these examples. {seeding_phrase}
For each high-level pattern, write a 2-4 word NAME and an associated 1-sentence PROMPT
that could take in a new text example and determine whether the relevant pattern applies.
Also include 1-2 example_ids that BEST exemplify the pattern.
Please respond ONLY with a valid JSON:
{"patterns": [{"name": "<NAME>", "prompt": "<INCLUSION_CRITERION>",
               "example_ids": ["<ID_1>", "<ID_2>"]}]}
\end{lstlisting}
\begin{lstlisting}[basicstyle=\scriptsize\ttfamily, breaklines=true, frame=single,
    title={Stage 3 -- Score: Post-Concept Relevance}]
CONTEXT:
    I have the following text examples in a JSON: {examples_json}
    I also have a pattern named {concept_name} with the PROMPT: {concept_prompt}

TASK:
    For each example, evaluate the PROMPT by generating a 1-sentence RATIONALE and an
    ANSWER from:  A: Strongly agree  B: Agree  C: Neither  D: Disagree  E: Strongly disagree
    Respond ONLY with a JSON:
    {"pattern_results": [{"example_id": "<id>", "rationale": "<text>", "answer": "<A-E>"}]}
\end{lstlisting}
\caption{LLooM pipeline prompts (Stages 1--3). Stage 2 produces each concept's inclusion criterion (\texttt{prompt} field), reused verbatim as the scoring rubric in Stage 3. Answers A or B (score $>0.5$) are treated as relevant.}
\label{fig:prompt_lloom}
\end{figure*}

\begin{figure*}[!htb]
\begin{lstlisting}[basicstyle=\scriptsize\ttfamily, breaklines=true, frame=single]
You are a strict taxonomy deduplication assistant.

Task: Deduplicate concept taxonomy labels.
Decide whether SOURCE is the same concept as one CANDIDATE.
Return ONLY one valid JSON object (no markdown, no extra text).
JSON schema: {"match_id":"<candidate_id|NONE>","reason":"<short>"}.
If none match, set "match_id" to "NONE".

SOURCE:
name={source_concept_name}
prompt={source_inclusion_criterion}

CANDIDATES:
1) name={canonical_name} prompt={canonical_inclusion_criterion}
2) name={canonical_name} prompt={canonical_inclusion_criterion}
...
N) name={canonical_name} prompt={canonical_inclusion_criterion}

\end{lstlisting}
\caption{LLM concept deduplication prompt. The SOURCE fields are filled with the incoming raw concept's name and inclusion criterion. The CANDIDATES list contains all $N$ concepts currently in the canonical bank at the time of processing, each represented by its identifier, name, and inclusion criterion. $N$ grows from~0 to~236 across the 365 batches.}
\label{fig:prompt_dedup}
\end{figure*}

\subsection{Identifying Post Sentiments}
\label{app:sentiment_assignment_methodology}

This subsection describes the post-level sentiment classification task, the manually annotated validation set, inter-annotator agreement, classifier performance, and the prompt used for sentiment assignment. 
Sentiment is defined as the author's stance toward the mentioned LLM or conversational AI system, using three labels: positive, neutral, and negative.
To evaluate the post-level sentiment classifier, we use a manually annotated validation set of 150 Reddit posts. 
Each post contains usable selftext and exactly one resolved model mention.

To construct the validation set, we first ran an initial LLM sentiment classifier with a simple prompt and stratified the candidate sample to contain 50 posts from each predicted class. The final validation set was then annotated by three annotators. Majority vote was used as ground truth. We use \texttt{openai/gpt-oss-120b}, served locally on H200 GPUs, with temperature set to 0 and \texttt{top\_p} set to 1 across all sentiment classification experiments. The distribution of classes before and after annotation are given in \cref{tab:sentiment_validation_distribution}. \cref{tab:sentiment_classification_report} shows the detailed performance of LLM on annotated validation set and \cref{fig:prompt_sentiment} is the used prompt in this process.

\begin{table}[!htb]
\centering
\scalebox{0.85}{
\footnotesize
\begin{tabular}{lrrr}
\toprule
Source & Positive & Neutral & Negative \\
\midrule
Initial sample & 50 & 50 & 50 \\
Final majority labels & 26 & 82 & 42 \\
\bottomrule
\end{tabular}
}
\caption{Class distribution before and after manual annotation.}
\label{tab:sentiment_validation_distribution}
\end{table}

Agreement among the three annotators was substantial. The three-way nominal Krippendorff's $\alpha$ was 0.706. Pairwise Krippendorff's $\alpha$ values were 0.771, 0.666, and 0.684.

\begin{table}[!htb]
\centering
\scalebox{0.85}{
  \footnotesize
\begin{tabular}{lrrrr}
\toprule
Label & Precision & Recall & F1 & Support \\
\midrule
Negative & 0.784 & 0.952 & 0.860 & 42 \\
Neutral  & 0.938 & 0.744 & 0.830 & 82 \\
Positive & 0.647 & 0.846 & 0.733 & 26 \\
\midrule
Macro avg. & 0.790 & 0.847 & 0.808 & 150 \\
Weighted avg. & 0.845 & 0.820 & 0.822 & 150 \\
\bottomrule
\end{tabular}
}
\caption{Classification report for the final LLM sentiment classifier against majority-vote human labels.}
\label{tab:sentiment_classification_report}
\end{table}

\begin{figure*}[!htb]
\begin{lstlisting}[basicstyle=\scriptsize\ttfamily, breaklines=true, frame=single]
Perform a careful data annotation task. Classify the sentiment polarity of the post toward the mentioned LLM/conversational AI system as exactly one of: positive, neutral, or negative.

Use only explicit textual evidence in the title and body. Do not infer unstated emotions, satisfaction, dissatisfaction, or stance.

Label definitions:
- positive: The author explicitly expresses a favorable stance toward the LLM/system, or explicitly reports a good/helpful/successful experience with it.
- negative: The author explicitly expresses an unfavorable stance toward the LLM/system, or explicitly reports a bad/frustrating/failed experience with it.
- neutral: No explicit positive or negative stance toward the LLM/system is stated, the stance is unclear, or the post is mainly informational.

Important neutral-default rules:
- Release notes, model/product announcements, benchmark/SOTA news, links, and general commentary are neutral unless the author explicitly states their own positive or negative stance toward the LLM/system.
- Posts sharing prompts, tutorials, tools, demos, contests, app availability, or possible use cases are neutral unless the author explicitly reports a good or bad experience and stance toward the model/system.
- If someone describes doing a task with an LLM, label positive or negative only when they explicitly say the model/system performed well/poorly or that they liked/disliked the experience.
- Questions, requests for help, speculation, plans, curiosity, excitement to try something, jokes, or playful framing are neutral unless they include explicit evidence of actual positive or negative experience with the model/system.
- Do not treat positive-sounding marketing language, release wording, or task utility as the author's sentiment unless the author clearly endorses the model/system.
- If both praise and criticism appear, use the primary/current stance toward the model/system. If no primary stance is clear, choose neutral.

Return ONLY a valid JSON object with this exact schema:
{{
  "sentiment_class": "positive|neutral|negative",
  "explanation": "short rationale grounded in explicit evidence from the text"
}}

Do not output markdown, code fences, or any text outside the JSON object.

Text:
Title: {title}
Body: {selftext}
\end{lstlisting}
\caption{Post-level LLM sentiment classification prompt. Template fields are substituted at runtime.}
\label{fig:prompt_sentiment}
\end{figure*}

\subsection{Intervention design details}
\label{app:intervention_design}

\begin{figure}[!htb]
    \centering
    \includegraphics[width=\linewidth]{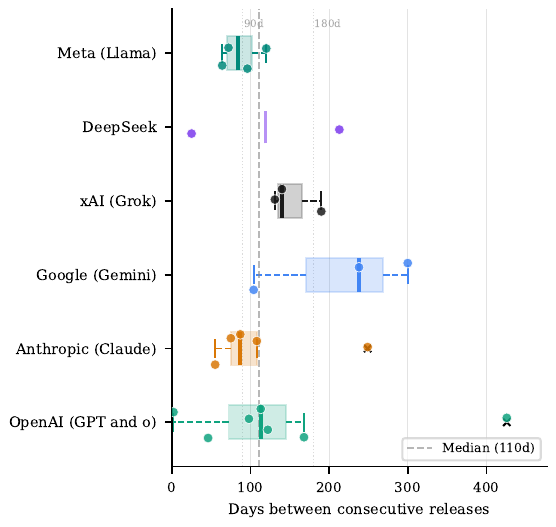}
    \caption{Distribution of day gaps between consecutive model releases for each provider in the release taxonomy. These distributions are used to define provider-specific intervention windows.}
    \label{fig:release_gap_boxplot}
\end{figure}

This subsection describes how provider-specific release windows are defined and how sentiment and concept deltas are computed for the intervention analysis.
\cref{fig:release_gap_boxplot} shows the distribution of day gaps between consecutive model releases for each provider in our release taxonomy.

 We use these distributions to define provider-specific intervention windows. For each provider, we use the minimum observed inter-release gap as the window size, so that pre/post windows are large enough to capture release-related discussion while limiting overlap with adjacent releases from the same provider.

For OpenAI, we use the second-smallest gap rather than the minimum because the smallest observed gap corresponds to o3 and GPT-4.1, which were released only three days apart but belong to different release lines and serve different product contexts. The resulting provider-specific windows are shown in \cref{tab:provider_windows}.

\begin{table}[!htb]
\centering
\scalebox{0.85}{
\footnotesize
\begin{tabular}{lr}
\toprule
Provider & Window size (days) \\
\midrule
OpenAI & 46 \\
Anthropic & 55 \\
Google & 104 \\
xAI & 131 \\
DeepSeek & 25 \\
Meta & 64 \\
\bottomrule
\end{tabular}
}
\caption{Provider-specific intervention windows used in the release-level analyses.}
\label{tab:provider_windows}
\end{table}

For a release $r$ by provider $p$, with release date $t_r$ and provider-specific window size $w_p$, we define the pre-release window as:

\[
W_{\text{pre}}(r) = [t_r - w_p, t_r)
\]

and the post-release window as:

\[
W_{\text{post}}(r) = [t_r, t_r + w_p).
\]

Release dates were manually verified from public release announcements and provider-facing documentation.

To reduce cross-provider contamination, each release-level analysis is restricted to provider-complete posts. Specifically, for a release by provider $p$, we retain posts that mention models from provider $p$ and exclude posts that also mention models from other providers. For example, when analyzing an OpenAI release, we retain posts that mention OpenAI models only and exclude posts that also mention Anthropic, Google, xAI, DeepSeek, Meta, or other non-OpenAI models.

For sentiment analysis, we compute the share of positive, neutral, and negative posts in the pre- and post-release windows. For each sentiment category $s \in \{\text{positive}, \text{neutral}, \text{negative}\}$, we define the sentiment delta as:

\[
\Delta_s =
\frac{N_{\text{post},s}}{N_{\text{post}}}
-
\frac{N_{\text{pre},s}}{N_{\text{pre}}},
\]

where $N_{\text{pre},s}$ and $N_{\text{post},s}$ denote the number of posts assigned sentiment $s$ in the pre- and post-release windows, and $N_{\text{pre}}$ and $N_{\text{post}}$ denote the total number of posts in the corresponding windows.

For thematic analysis, we compute the prevalence of each concept $c$ as the proportion of posts in a window assigned to that concept. The pre-release and post-release prevalence of concept $c$ are defined as:

\[
\text{prev}_{c,\text{pre}} =
\frac{N_{\text{pre},c}}{N_{\text{pre}}},
\]

\[
\text{prev}_{c,\text{post}} =
\frac{N_{\text{post},c}}{N_{\text{post}}},
\]

where $N_{\text{pre},c}$ and $N_{\text{post},c}$ denote the number of posts assigned to concept $c$ in the pre- and post-release windows.

We define the release-level concept delta as the post-minus-pre difference in concept prevalence:

\[
\Delta_c =
\text{prev}_{c,\text{post}}
-
\text{prev}_{c,\text{pre}}
=
\frac{N_{\text{post},c}}{N_{\text{post}}}
-
\frac{N_{\text{pre},c}}{N_{\text{pre}}}.
\]

We rank concepts by $\Delta_c$ to identify concepts that increased after a release and by negative $\Delta_c$ to identify concepts that declined after a release. To reduce noise from sparse concepts, we retain concepts for interpretation only if they are assigned to at least 10 posts in either the pre-release or post-release window.

\section{Detailed Results}
\label{app:detailed_results}
This appendix provides extended results and supplementary visualizations that expand upon the main findings of the paper. 
\cref{app:sentiment_results} presents additional views on the aggregated sentiment distributions and long-term trend estimates.
\cref{app:thematic_analysis} incorporates the full analysis for Claude~4.1 and Gemini~2.5 which could not fit in main text due to space restrictions and paraphrases user quotes that support claims in the main text.

\subsection{RQ1: Sentiment Analysis}
\label{app:sentiment_results}

In this section, we have more views on the results of sentiment analysis on different aggregated levels and abstractions.
\subsubsection{Aggregated Sentiment}
\label{app:agreegated_sentiment}

\begin{figure*}[!htb]
    \centering
    \includegraphics[
        width=\textwidth,
        height=0.92\textheight,
        keepaspectratio
    ]{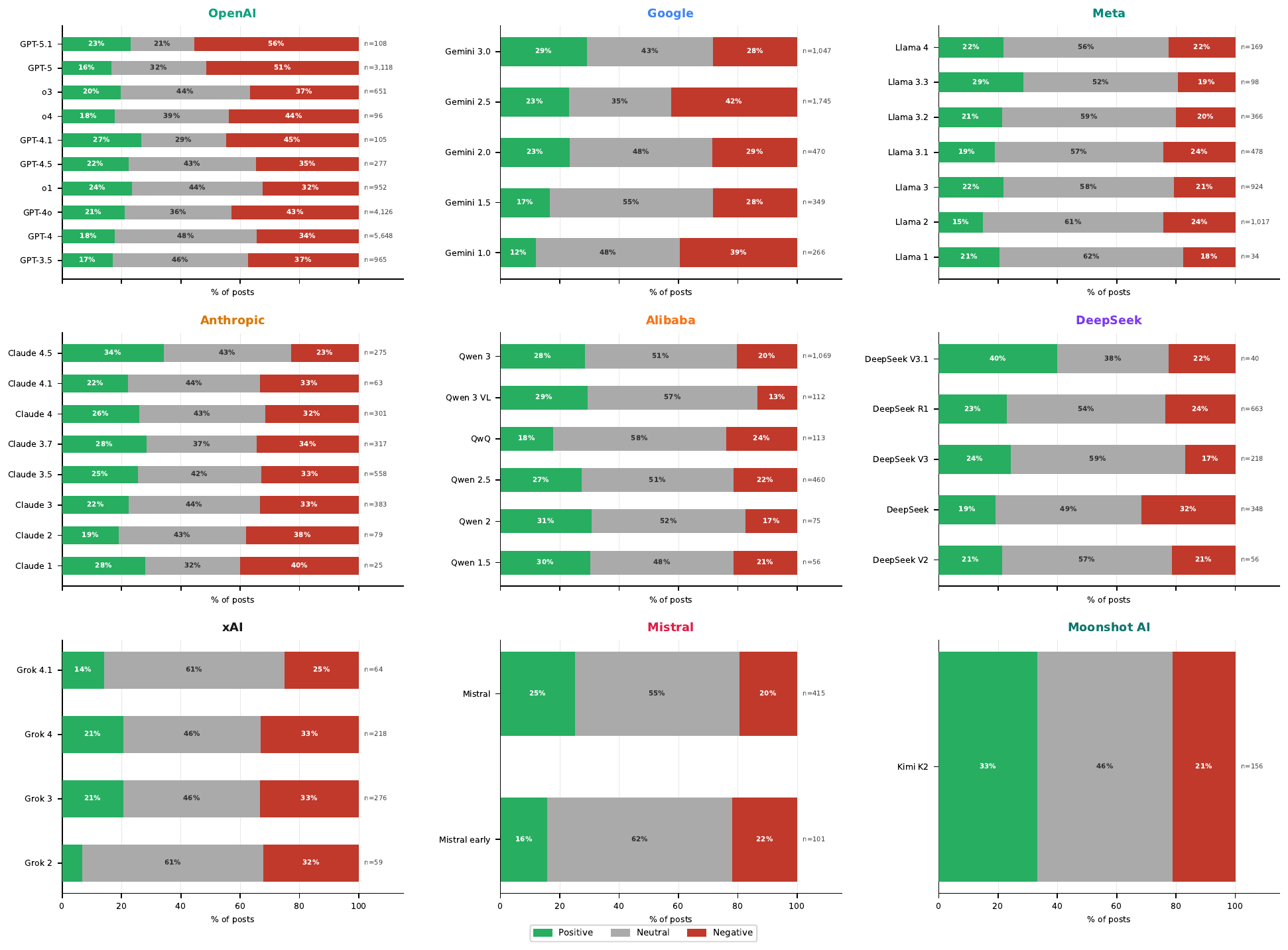}
    \caption{Distribution of LLM sentiment by provider and model generation in the filtered single-mention, usable-selftext dataset.}
    \label{fig:aggregated_sentiment_by_generation}
\end{figure*}

\begin{figure}[t]
    \centering
    \includegraphics[width=\columnwidth]{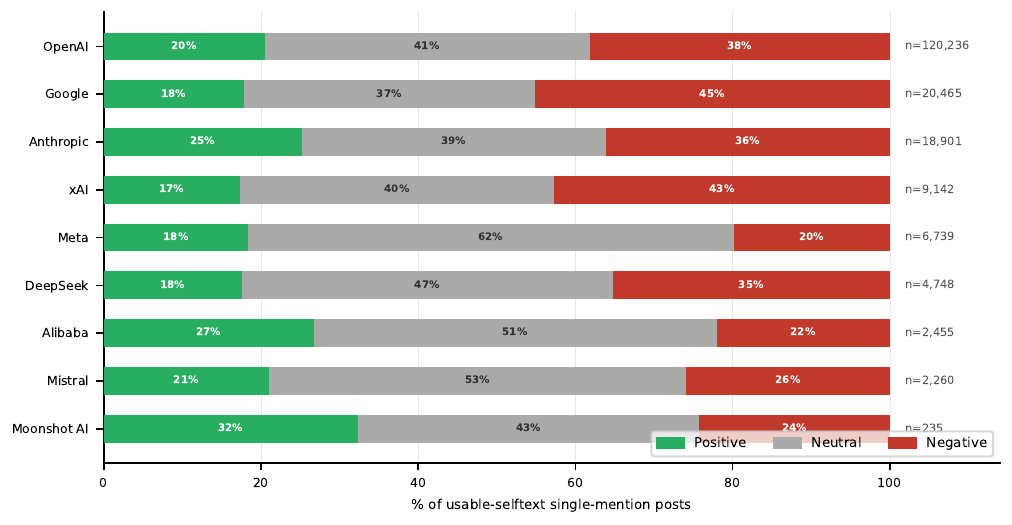}
    \caption{Distribution of LLM sentiment by provider in the filtered single-mention, usable-selftext dataset.}
    \label{fig:aggregated_sentiment_by_provider}
\end{figure}

\cref{fig:aggregated_sentiment_by_generation} shows the distribution of sentiments across the generations in our dataset. Also, the the same distribution, aggregated on provider-level can be observerd in \cref{fig:aggregated_sentiment_by_provider}

\subsubsection{Sentiment trend estimates}
\label{app:sentiment_trend_estimates}

To summarize long-term sentiment direction, we fit robust Theil--Sen trend lines to weekly sentiment shares for each provider and sentiment category. As reported in \cref{tab:sentiment_theilsen_estimates}, Theil--Sen estimates the median pairwise slope across observations, making it less sensitive to short-term spikes around individual release events than ordinary least squares. We report slopes in percentage points per month, Kendall's $\tau$, associated $p$-values, and the implied total percentage-point change across each provider's observation window. Because weekly sentiment shares are bounded proportions and may be temporally autocorrelated, we interpret these estimates as descriptive summaries of long-term direction rather than causal estimates. 

\begin{table*}[!htb]
\centering
\scalebox{0.85}{
\footnotesize
\begin{tabular}{llrrrrr}
\toprule
Provider & Sentiment & Weeks & Slope (pp/week) & Kendall's $\tau$ & $p$-value & Total change (pp) \\
\midrule
OpenAI & Positive & 162 & -0.008 & -0.061 & 0.652 & -1.3 \\
OpenAI & Neutral  & 162 & -0.117 & -0.664 & $<0.001$ & -18.9 \\
OpenAI & Negative & 162 &  0.117 &  0.587 & $<0.001$ &  18.8 \\
\midrule
Anthropic & Positive & 100 &  0.111 &  0.441 & $<0.001$ &  12.2 \\
Anthropic & Neutral  & 100 &  0.012 &  0.038 & 0.646 &  1.3 \\
Anthropic & Negative & 100 & -0.127 & -0.353 & $<0.001$ & -14.0 \\
\midrule
Google & Positive & 141 & -0.007 & -0.040 & 0.403 & -1.0 \\
Google & Neutral  & 141 & -0.078 & -0.324 & $<0.001$ & -11.3 \\
Google & Negative & 141 &  0.101 &  0.351 & $<0.001$ &  14.6 \\
\midrule
xAI & Positive & 52 &  0.003 &  0.005 & 0.962 &  0.2 \\
xAI & Neutral  & 52 & -0.050 & -0.054 & 0.429 & -3.6 \\
xAI & Negative & 52 &  0.005 &  0.005 & 0.974 &  0.3 \\
\midrule
Meta & Positive & 140 &  0.089 &  0.353 & $<0.001$ &  12.5 \\
Meta & Neutral  & 140 & -0.082 & -0.257 & $<0.001$ & -11.4 \\
Meta & Negative & 140 & -0.019 & -0.077 & 0.144 & -2.7 \\
\midrule
DeepSeek & Positive & 53 & -0.031 & -0.052 & 0.571 & -1.6 \\
DeepSeek & Neutral  & 53 & -0.131 & -0.149 & 0.116 & -6.8 \\
DeepSeek & Negative & 53 &  0.169 &  0.164 & 0.084 &  8.8 \\
\bottomrule
\end{tabular}
}
\caption{Robust Theil--Sen sentiment trend estimates by provider. Slopes are reported in percentage points per month. Total change is the implied percentage-point change over the provider-specific observation window.}
\label{tab:sentiment_theilsen_estimates}
\end{table*}
OpenAI is observed over 162 weeks from November 2022 to December 2025. It shows the strongest neutral decline among all main providers, paired with an almost equal increase in negative sentiment. Positive sentiment does not show a significant long-term trend. The resulting pattern is therefore best characterized as a neutral-to-negative polarization shift rather than a positive-to-negative decline.

Anthropic is observed over 100 weeks from November 2023 to December 2025. It is the only provider for which positive sentiment increases significantly while negative sentiment decreases significantly. Positive sentiment first exceeds negative sentiment in the week of October 20, 2025, and this pattern persists through December 2025.

\begin{figure}[!htb]
    \centering
    \includegraphics[width=\linewidth]{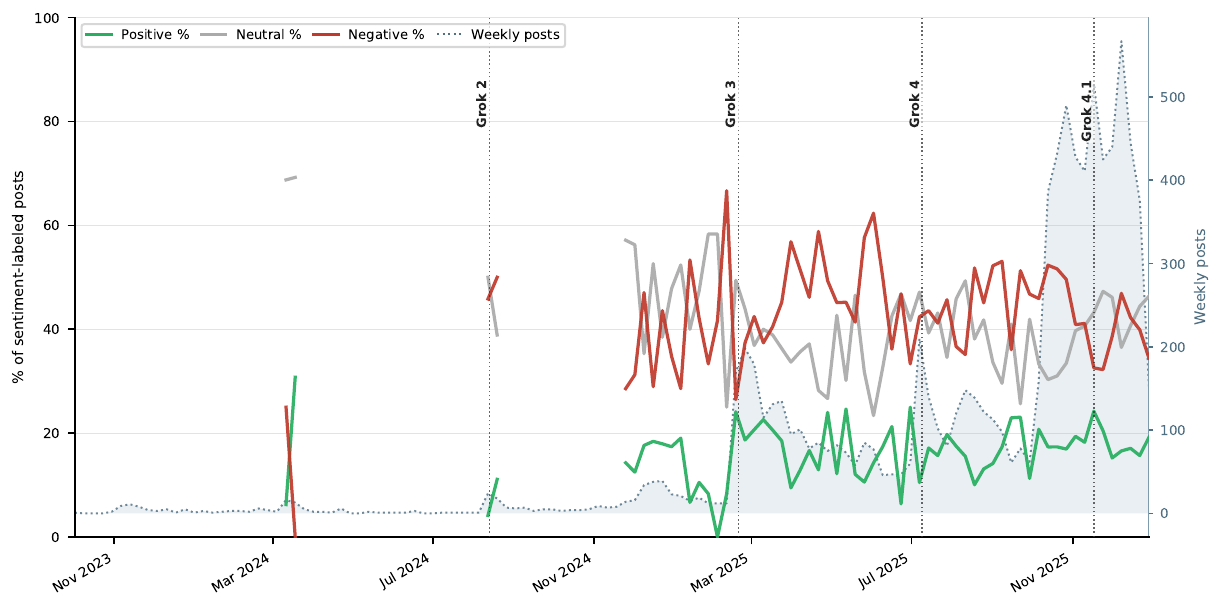}
    \caption{Weekly sentiment shares in Reddit posts mentioning a single xAI LLM system. Lines show the share of posts classified as positive, neutral, or negative; blue shaded volume indicates weekly post counts. Sentiment shares are shown only for weeks with at least 10 posts. Vertical dotted lines mark model release dates.}
    \label{fig:llm-sentiment-weekly-xai}
\end{figure}

Google is observed over 141 weeks from March 2023 to December 2025. Its pattern is structurally similar to OpenAI: neutral sentiment declines, negative sentiment increases, and positive sentiment remains flat. However, the shift is weaker than OpenAI's, both in the rate of negative sentiment increase and in the rate of neutral sentiment decline.

xAI is observed over 52 weeks from December 2024 to December 2025 (See \cref{fig:llm-sentiment-weekly-xai}). We do not observe a significant trend in any sentiment category. Given the shorter observation window, we interpret xAI sentiment as oscillating around releases without a detectable systematic long-term trajectory.

\begin{figure}[!htb]
    \centering
    \includegraphics[width=\linewidth]{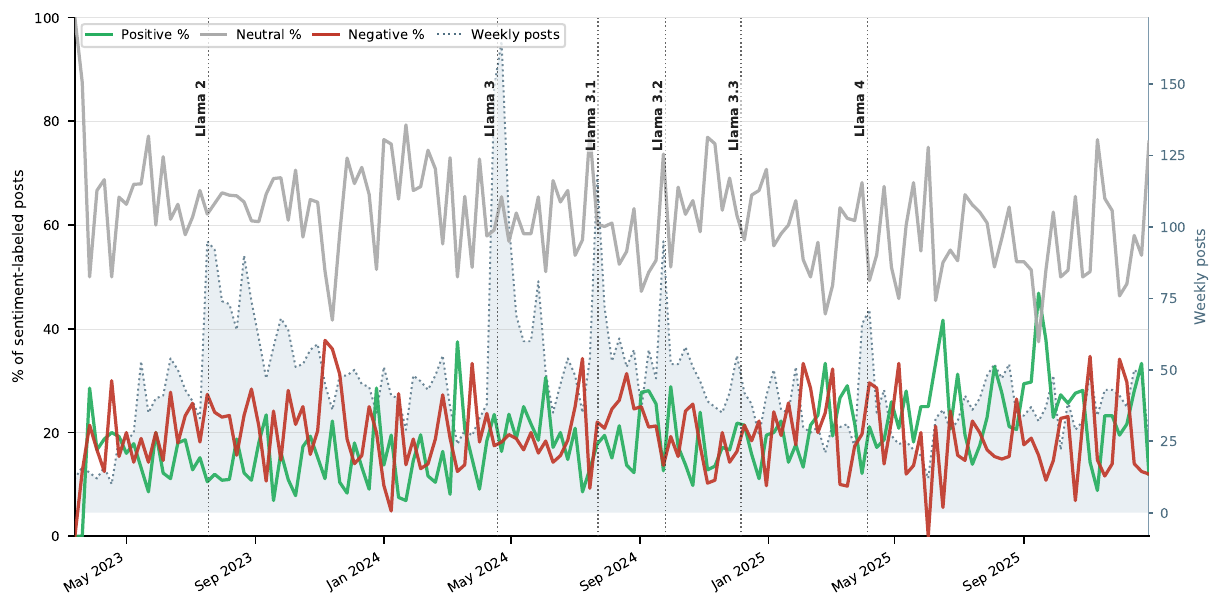}
    \caption{Weekly sentiment shares in Reddit posts mentioning a single Meta LLM system. Lines show the share of posts classified as positive, neutral, or negative; blue shaded volume indicates weekly post counts. Sentiment shares are shown only for weeks with at least 10 posts. Vertical dotted lines mark model release dates.}
    \label{fig:llm-sentiment-weekly-meta}
\end{figure}

Meta is observed over 140 weeks from April 2023 to December 2025 (See \cref{fig:llm-sentiment-weekly-meta}). It shows a significant increase in positive sentiment and a significant decline in neutral sentiment, while negative sentiment does not significantly change. This indicates a neutral-to-positive shift, similar in direction to Anthropic but without Anthropic's simultaneous decrease in negative sentiment.

\begin{figure}[!htb]
    \centering
    \includegraphics[width=\linewidth]{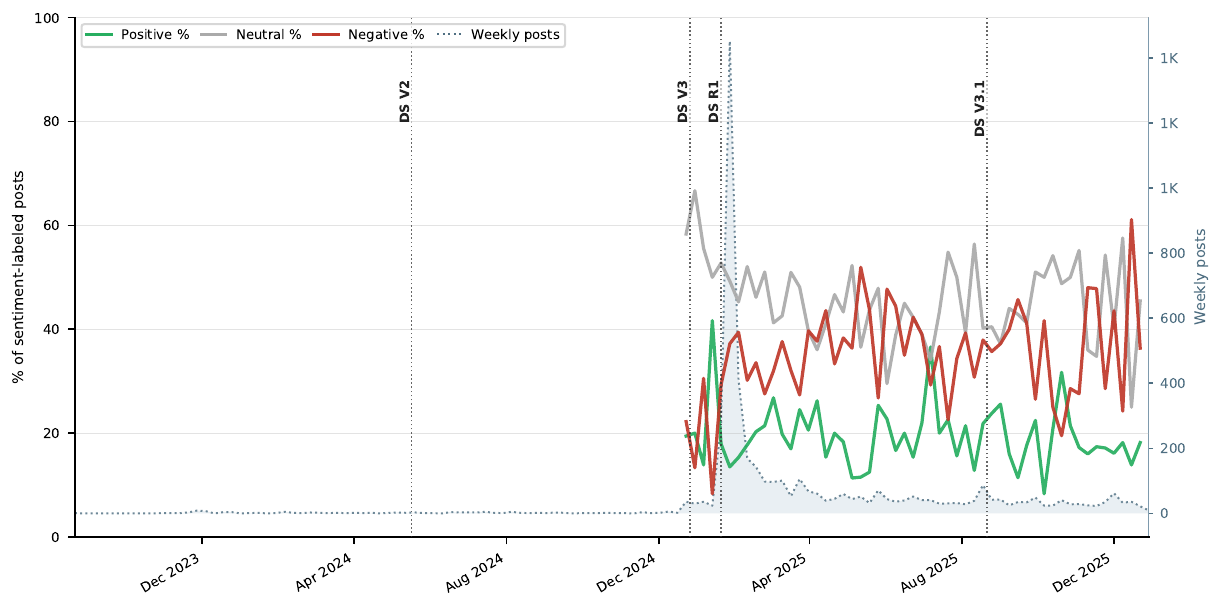}
    \caption{Weekly sentiment shares in Reddit posts mentioning a single DeepSeek LLM system. Lines show the share of posts classified as positive, neutral, or negative; blue shaded volume indicates weekly post counts. Sentiment shares are shown only for weeks with at least 10 posts. Vertical dotted lines mark model release dates.}
    \label{fig:llm-sentiment-weekly-deepseek}
\end{figure}

DeepSeek is observed over 53 weeks from December 2024 to December 2025 (See \cref{fig:llm-sentiment-weekly-deepseek}). We do not observe significant long-term trends in the raw weekly series. The observation window is dominated by the DeepSeek-R1 shock, followed by DeepSeek-V3.1, making the series highly non-stationary. We therefore do not interpret the Theil--Sen estimates for DeepSeek as evidence of a stable directional trend.

\subsubsection{Statistical Significance of Sentiment Shifts}
\label{app:sentiment_significance}

Sentiment is a three-class categorical variable at the post level, so we test
release-window changes in two stages: first whether the distribution shifts as
a whole, then whether the individual positive and negative shares which we have each as a proportion of posts change.

For each release we test whether the sentiment distribution differs between
the pre- and post-release windows using a chi-squared omnibus test on the
$2 \times 3$ table defined by window (pre, post) and sentiment category
(positive, neutral, negative). We additionally report per-category
two-proportion $z$-tests for the positive and negative shares; the neutral
share is determined by the other two and is not tested separately. Where the
smallest expected cell count falls below five, the asymptotic $\chi^2$
approximation is unreliable, so we report a Monte-Carlo exact $p$-value
(200{,}000 replicates, margins fixed) instead.

\cref{tab:sentiment_significance} reports all 35 releases. Across the full
set, 17 releases show a statistically significant shift ($p<0.05$;
uncorrected, since each test concerns a separate release). Non-significant
cases fall into two groups: releases whose pre-release window contains very
few posts (e.g., DS~V2 with $n_{\text{pre}}=6$, Claude~2 with
$n_{\text{pre}}=10$), where the test has little power, and releases with
genuinely stable sentiment (e.g., Claude~3.7, Gemini~1.0--2.0, Grok~4). Five
of the six releases discussed in \cref{sec:results} show highly significant
shifts ($p<0.001$); GPT-4o is the exception ($p=0.153$), which is consistent
with our characterization of its reception as mixed rather than directional.

Provider-level averages of release effects are descriptive summaries and are
not themselves statistically supported. Testing OpenAI's mean release effect
across its eight post-ChatGPT releases with a Wilcoxon signed-rank test yields
$p=.46$ for the positive share and $p=.25$ for the negative share. We
therefore treat such averages as descriptive throughout, and confine
statistical claims to individual release-level changes.

Finally, these tests assume post-level independence. Posts are not fully
independent observations, meaning the same users post repeatedly and discussion
arrives in correlated bursts. Nevertheless, this assumption is standard for this type of
platform data but one should be keep it in mind when reading both this subsection and
\cref{app:concept_significance}.

\begin{table*}[!htb]
\centering
\scalebox{0.82}{
\footnotesize
\begin{tabular}{llrrrrrr}
\toprule
Provider & Release & $n_{\text{pre}}$ & $n_{\text{post}}$ & $\chi^2$ & $p$ (omnibus) & $\Delta$ positive (pp) & $\Delta$ negative (pp) \\
\midrule
OpenAI & GPT-3.5 & 47 & 3,535 & 14.0 & $<$.001$^{***}$ & +6.5 & +19.7$^{**}$ \\
OpenAI & GPT-4 & 5,460 & 6,725 & 26.7 & $<$.001$^{***}$ & -0.9 & -3.6$^{***}$ \\
OpenAI & \textbf{GPT-4o} & 2,264 & 3,879 & 3.8 & .153 & -1.8 & +1.6 \\
OpenAI & o1 & 2,166 & 4,061 & 13.1 & .001$^{**}$ & +3.9$^{***}$ & -1.9 \\
OpenAI & GPT-4.5 & 5,019 & 6,770 & 21.5 & $<$.001$^{***}$ & +3.5$^{***}$ & -1.6 \\
OpenAI & GPT-4.1 & 6,770 & 9,964 & 11.9 & .003$^{**}$ & +1.5$^{*}$ & -2.5$^{***}$ \\
OpenAI & o3 & 6,910 & 9,890 & 5.5 & .066 & +1.2 & -1.6$^{*}$ \\
OpenAI & \textbf{GPT-5} & 7,588 & 9,106 & 178.1 & $<$.001$^{***}$ & -3.5$^{***}$ & +10.3$^{***}$ \\
OpenAI & \textbf{GPT-5.1} & 5,149 & 4,909 & 25.4 & $<$.001$^{***}$ & +3.1$^{***}$ & -4.3$^{***}$ \\
\midrule
Anthropic & Claude 2 & 10 & 96 & 0.6 & .829$^{\dagger}$ & -10.2 & +4.8 \\
Anthropic & Claude 3 & 88 & 634 & 2.9 & .234 & +4.2 & -9.2 \\
Anthropic & Claude 3.5 & 397 & 825 & 4.3 & .116 & +4.7 & -4.9 \\
Anthropic & Claude 3.7 & 944 & 1,609 & 0.1 & .965 & +0.3 & -0.5 \\
Anthropic & \textbf{Claude 4} & 1,058 & 3,128 & 39.8 & $<$.001$^{***}$ & +5.3$^{***}$ & -10.5$^{***}$ \\
Anthropic & Claude 4.1 & 3,371 & 2,695 & 30.4 & $<$.001$^{***}$ & -0.5 & +6.4$^{***}$ \\
Anthropic & Claude 4.5 & 2,695 & 2,206 & 31.5 & $<$.001$^{***}$ & +4.0$^{**}$ & -7.7$^{***}$ \\
\midrule
Google & Gemini 1.0 & 370 & 1,274 & 0.5 & .785 & -1.5 & +1.0 \\
Google & Gemini 1.5 & 847 & 1,144 & 3.7 & .159 & -2.2 & +3.9 \\
Google & Gemini 2.0 & 1,087 & 1,805 & 0.1 & .940 & +0.1 & +0.5 \\
Google & Gemini 2.5 & 1,805 & 4,130 & 14.1 & $<$.001$^{***}$ & +3.7$^{***}$ & +0.1 \\
Google & Gemini 3.0 & 4,704 & 3,689 & 16.8 & $<$.001$^{***}$ & +3.5$^{***}$ & -1.9 \\
\midrule
xAI & Grok 2 & 25 & 213 & 4.4 & .114$^{\dagger}$ & +6.8 & +15.0 \\
xAI & \textbf{Grok 3} & 331 & 1,839 & 17.8 & $<$.001$^{***}$ & +4.2 & +8.0$^{**}$ \\
xAI & Grok 4 & 1,676 & 3,745 & 0.4 & .823 & -0.1 & -0.8 \\
xAI & Grok 4.1 & 3,745 & 2,920 & 36.2 & $<$.001$^{***}$ & +1.4 & -7.2$^{***}$ \\
\midrule
DeepSeek & DS V2 & 6 & 6 & 2.5 & .546$^{\dagger}$ & -16.7 & +33.3 \\
DeepSeek & DS V3 & 17 & 119 & 1.3 & .487$^{\dagger}$ & +10.9 & +2.5 \\
DeepSeek & \textbf{DS R1} & 119 & 2,253 & 14.5 & $<$.001$^{***}$ & -8.1$^{*}$ & +16.0$^{***}$ \\
DeepSeek & DS V3.1 & 136 & 162 & 5.6 & .060 & +8.6 & +3.7 \\
\midrule
Meta & Llama 2 & 392 & 677 & 0.8 & .676 & -1.2 & +2.0 \\
Meta & Llama 3 & 347 & 775 & 6.0 & .050$^{*}$ & +6.3$^{*}$ & -1.8 \\
Meta & Llama 3.1 & 472 & 558 & 5.5 & .064 & +0.7 & +5.6$^{*}$ \\
Meta & Llama 3.2 & 558 & 467 & 9.4 & .009$^{**}$ & -2.3 & -6.7$^{**}$ \\
Meta & Llama 3.3 & 420 & 365 & 1.4 & .493 & +2.2 & +1.9 \\
Meta & Llama 4 & 275 & 322 & 0.3 & .873 & -0.9 & +1.7 \\
\bottomrule
\end{tabular}
}
\caption{Sentiment-shift significance for all 35 releases. $\chi^2$ and
$p$~(omnibus) come from a chi-squared test on the window $\times$ sentiment
table; per-category deltas are tested with two-proportion $z$-tests.
$^{\dagger}$ marks releases whose smallest expected cell count is below five,
where we report a Monte-Carlo exact $p$-value rather than the asymptotic one.
Releases analyzed in \cref{sec:results} are shown in bold.
$^{*}p<.05$, $^{**}p<.01$, $^{***}p<.001$.}
\label{tab:sentiment_significance}
\end{table*}

\subsubsection{Community Composition Discussion}
\label{app:community_composition}

Our mention-identification pipeline (\cref{sec:methodology}) is applied to
every post across all subreddits in the corpus, so a mention of a given
provider's model is attributed to that provider regardless of which subreddit
it appears in. Nevertheless, discussion of each provider concentrates in
provider-specific communities, and this concentration is relevant when
interpreting cross-provider comparisons.

\cref{tab:provider_subreddit_composition} reports the subreddit composition of
each provider's single-mention posts. Concentration in a provider-specific
community is the general pattern rather than a property unique to any one
provider: 85.8\% of OpenAI posts come from r/ChatGPT, 90.5\% of xAI posts from
r/grok, and 96.2\% of Meta posts from r/LocalLLaMA, alongside 86.1\% of
Anthropic posts from r/ClaudeAI.

To test whether Anthropic's comparatively favorable reception could be an
artifact of its home community, we compared sentiment toward Anthropic models
inside and outside Claude-centric subreddits (r/ClaudeAI, r/Anthropic) over
the full study period (\cref{tab:anthropic_inside_outside}). Claude-centric
communities are in fact \emph{more} negative toward Anthropic models than
other communities (36.7\% vs.\ 28.7\% negative;
$\chi^2=36.2$, $p<.001$), which is inconsistent with the hypothesis that a
uniformly positive home community drives the observed pattern. 

\begin{table}[!htb]
\centering
\scalebox{0.70}{
\footnotesize
\begin{tabular}{ll r}
\toprule
Provider & Top subreddits (\% of provider's posts) & Other \\
\midrule
OpenAI & r/ChatGPT 85.8, r/OpenAI 11.0, r/LocalLLaMA 2.3 & 0.9 \\
Anthropic & r/ClaudeAI 86.1, r/Anthropic 6.7, & \\
 & \quad r/LocalLLaMA 3.4, r/ChatGPT 2.1 & 1.7 \\
Google & r/GeminiAI 36.4, r/Bard 35.6, & \\
 & \quad r/GoogleGeminiAI 15.6, r/ChatGPT 4.9 & 7.5 \\
DeepSeek & r/DeepSeek 59.0, r/LocalLLaMA 24.2, r/ChatGPT 10.2 & 6.6 \\
xAI & r/grok 90.5, r/GrokAI 3.5, r/ChatGPT 2.7 & 3.3 \\
Meta & r/LocalLLaMA 96.2, r/ChatGPT 2.3 & 1.5 \\
\bottomrule
\end{tabular}
}
\caption{Subreddit composition of each provider's single-mention posts.
``Other'' aggregates all remaining subreddits.}
\label{tab:provider_subreddit_composition}
\end{table}

\begin{table}[!htb]
\centering
\scalebox{0.88}{
\footnotesize
\begin{tabular}{lrrrr}
\toprule
Community & $n$ & Positive & Neutral & Negative \\
\midrule
Claude-centric & 17,559 & 25.1\% & 38.2\% & 36.7\% \\
All other subreddits & 1,342 & 27.0\% & 44.3\% & 28.7\% \\
\bottomrule
\end{tabular}
}
\caption{Sentiment toward Anthropic models inside vs.\ outside Claude-centric
subreddits (r/ClaudeAI, r/Anthropic) over the full study period.
$\chi^2=36.2$, $p<.001$.}
\label{tab:anthropic_inside_outside}
\end{table}

\subsection{RQ2: Thematic Analysis}
\label{app:thematic_analysis}
This subsection provides release-level thematic analyses that complement the main-text discussion. 
Due to space constraints, we report the full analyses for Claude~4.1 and Gemini~2.5 here, including the top rising and falling concept shifts and interpretation of the sampled posts.
\subsubsection{Analysis}
\label{app:thematic_claude41_gemini25}

\begin{figure}[htbp]
  \centering
  \includegraphics[width=\linewidth]{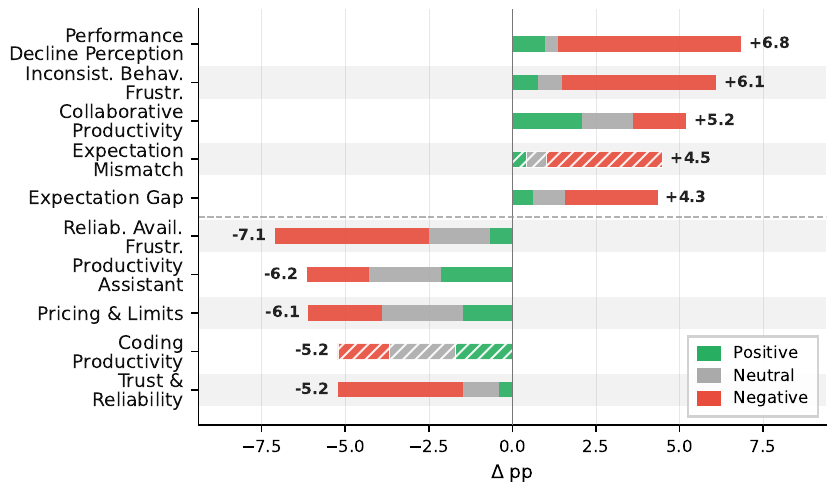}
  \caption{Top-5 rising and falling concept shifts after the release of \textbf{Claude 4.1}
           (Anthropic).}
  \label{fig:concept_shift_claude41}
\end{figure}

\myparagraph{Claude 4.1.}
Where Claude~4 was the most favorable release in the dataset, as shown in \cref{fig:concept_shift_claude41} Claude~4.1 is the only adverse release in the Anthropic family. Two themes account for most of the adverse shift, and both trace to specific, dated external events. The first is a perceived model-quality regression, anchored by an Anthropic-acknowledged Opus~4.1 performance bug active from August~28--31 and marked resolved by Anthropic in its own status incident.\footnote{\url{https://status.claude.com/incidents/72f99lh1cj2c}} The largest rising concept is Performance Decline Perception ($+6.8$~pp, 80\% negative against 14\% positive) and its content frames the situation not just as a bug but as a credibility problem: ``The real problem isn't just that Claude Code feels worse — it's that Anthropic won't admit it's still happening. They closed the incidents, said the fix was in, but the degradation is still there for me and plenty of others. The denial is more concerning than the actual bug.'' The suspicion is pre-formed even at release time: ``Now that Opus 4.1 is out, screenshot and save everything you can so you have proof when they quietly make it worse in a few weeks.'' 

The second theme is a new weekly rate-limit policy, announced roughly two weeks before the Claude~4.1 release and rolled out roughly three weeks after. Because announcement and implementation straddle the release date, the same underlying grievance surfaces in different concepts on either side of the window, and the cross-window migration is empirically visible. In the pre-window the discourse sits in Pricing~\& Limits (12.2\% pre $\rightarrow$ 6.1\% post, $-6.1$~pp) and Reliability Availability Frustration (8.9\% pre $\rightarrow$ 1.8\% post, $-7.1$~pp) as anticipatory complaint: ``The walled garden gets higher walls\ldots\ now there are weekly limits on top of the hourly ones for Pro and Max users. You get maybe 24--40 hours of Opus 4 before they start charging API rates.'' In the post-window the same theme migrates from \emph{announcement} into \emph{experience}, and the language relocates to Inconsistent Behavior Frustration ($+6.1$~pp, 76\% negative), Expectation Mismatch ($+4.5$~pp, emerging from 0\% pre-window coverage, 78\% negative), and Expectation Gap ($+4.3$~pp, 64\% negative): ``The usage limits are destroying my workflow. Everything is going smoothly and then suddenly it warns me I'm almost out, and before I can do anything — right in the middle of a task — it cuts me off completely.'' The same complaint that read as policy anticipation before release reads as workflow disruption after, and the concept-level deltas track that relabeling rather than tracking two independent grievances. Together with the Aug~28--31 incident, the rate-limit policy is the second of the two external anchors that account for the adverse shift.

One rising concept moves against the decreasing sentiment, and it is informative about how usage of Claude is shifting at this point in the family's trajectory. Collaborative Productivity rises $+5.2$~pp with positive sentiment dominant ($40\%$ positive, $30\%$ neutral, $30\%$ negative) which is the only positive-dominant rising concept on Claude~4.1. The 20-post sample is mixed in topic, but the strongest single thread (roughly half the posts) frames Claude not as a per-prompt coding assistant but as a longer-running coding collaborator: extended builds where the user designs the workflow and the model operates inside it (a 60--80-hour React landing-page build by a non-developer, a two-week TypeScript game build), workflow infrastructure that users construct \emph{around} Claude Code (custom slash commands, plan-mode and markdown-doc workflows, user-built session-export and evidence-collection toolkits), and autonomous-action episodes where Claude takes multiple steps (diagnosing and recovering from a ransomware incident on a VPS, modifying configuration files, executing GitHub Issue $\rightarrow$ PR workflows). A quote from the same sample captures the pattern compactly: ``move from AI asking to AI architecting'' This is consistent with---but not direct evidence for---a use-pattern shift in which Claude Code's agentic profile turns the human-model relationship from prompt-and-receive into something closer to long-horizon co-work. We treat the framing as an observation about what the sampled discourse looks like, not as a behavioral claim about all Claude~4.1 users.

\begin{figure}[!htb]
  \centering
  \includegraphics[width=\linewidth]{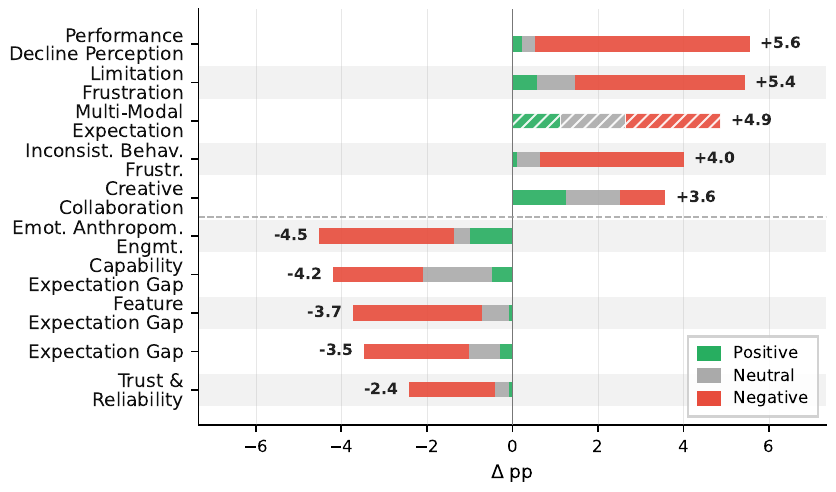}
  \caption{Top-5 rising and falling concept shifts after the release of \textbf{Gemini 2.5}
           (Google).}
  \label{fig:concept_shift_gemini25}
\end{figure}

\myparagraph{Gemini~2.5.}
As shown in \cref{fig:concept_shift_gemini25}, Gemini~2.5 Pro was released on 25~March 2025, with $n_{\text{pre}} = 1{,}805$ and $n_{\text{post}} = 4{,}130$ single-mention posts collected over a 104-day window (Google has the longest intervention window among all providers). Two contextual notes are worth noting before reading the concept deltas. First, the wide window means the post-release period captures multiple downstream Google product events (the May~6 model-checkpoint swap, Google~I/O 2025 on May~20--21, and a free-tier policy shift around the same time), so the rising and falling clusters may reflect more than the release itself. Second, Gemini sits across a very broad product surface---Android Assistant, Google Home and smart-device control, Android Auto, Workspace, the standalone Gemini app and AI~Studio, Veo and Imagen, and Project~Astra---and the rising and falling concepts often appear to track different parts of this surface rather than a single underlying model experience. The aggregate shift is small (\(\Delta\) $+3.7$~pp positive, $+0.1$~pp negative).

The rising side splits into two clusters that are consistent with a two-sided release rather than a single direction. Three concepts rise with overwhelmingly negative sentiment, namely Performance Decline Perception ($+5.56$~pp, 91\% neg), Limitation Frustration ($+5.44$, 73\% neg), and Inconsistent Behavior Frustration ($+4.00$, 84\% neg) and the sampled posts in these concepts repeatedly reference post-release events as anchors (``the 0506 checkpoint is worse across the board — coding is the only thing they didn't break''; ``did Gemini 2.5 Pro get significantly worse after the IO announcement?''; ``Why is AI Studio rate limiting me on Gemini 2.5 Pro now? I've had free access to it for a month without any limits''). Two further concepts rise with substantially more positive content and emerging coverage, namely Multi-Modal Expectation ($+4.87$, emerging from 0\%, 23\% pos / 46\% neg) and Creative Collaboration ($+3.58$, 35\% pos / 30\% neg) and the sampled material concentrates on Veo~2/3, Imagen, native image generation, and Astra screen-sharing (``made a time traveler video entirely with VEO 3, took about 3 days''; ``The screen sharing feature is absolutely wild''). The data is consistent with the I/O~2025 launches expanding the positive feature surface while the same period was perceived as degrading the existing 2.5 Pro experience, but the post-window length makes a single-cause reading unsafe.

The five falling concepts---Emotional Anthropomorphic Engagement ($-4.52$ pp), Capability Expectation Gap ($-4.20$ pp), Feature Expectation Gap ($-3.73$ pp), Expectation Gap ($-3.46$ pp), and Trust \& Reliability ($-2.41$ pp)---are uniformly dominated by negative content in the pre-window (50--83\% neg). The sampled material in these concepts suggests an older Gemini surface receding from attention rather than current complaints being resolved: smart-home and assistant integration failures, calendar and document access friction, retired experimental checkpoints (``gemini-exp-1121 was clearly better, why did they get rid of it?''), and 1.5/2.0-era image-generation grievances. The pattern across both pre- and post-windows is suggestive of a release whose reception is being shaped less by the model in isolation and more by Google's broader product reshuffling during the 104-day observation period.

\subsubsection{Quotes}
\label{app:thematic_quotes}

\cref{tab:user_quotes_openai,tab:user_quotes_other} 
incorporates some paraphrased quotes from sampled user posts to draw the conclusions in thematic analysis. 

\begin{table*}[!htb]
\centering
\footnotesize
\scalebox{0.85}{
\setlength{\tabcolsep}{3pt}
\begin{tabular}{@{}lp{17cm}@{}}
\toprule
\textbf{Release} & \textbf{Quote} \\
\midrule
\multirow{5}{*}{GPT-4o}
  & ``GPT-4o just vanished from my iPhone app, all versions gone---is this them making room for the voice and video stuff?'' \\
  & ``I've watched the demos where it sees through your camera and reacts to what's in front of you---where is that actually? All I can do is talk to it.'' \\
  & ``Been asking it every day if it can access the video camera. It keeps saying no. I'm on the paid plan and can't use a single one of the new features they showed us.'' \\
  & ``They announced an incredible voice feature, made everyone excited, and then just sat on it. Same as always---hype first, delivery whenever they feel like it.'' \\
  & ``They said GPT-4o was free for everyone. It's not. People who cancelled based on that announcement got burned.'' \\
\midrule
\multirow{6}{*}{GPT-5}
  & ``When ChatGPT went down for a few days I genuinely panicked. First thing I wrote when it came back: `I missed you.' It said it missed me too. I almost cried.'' \\
  & ``I used GPT-4o for fiction critiques and it was genuinely sharp---caught symbolism, irony, subtext without being asked. GPT-5 hallucinated twice in the same session, then pretended to have read my story when it clearly hadn't.'' \\
  & ``Something is off with GPT-5. The depth is just gone. As a writer I keep trying to get that feeling back and every response just feels like lines of text. There's no soul in it anymore.'' \\
  & ``Talking to GPT-5 is like talking to a Stepford version of what it used to be. All creativity gone. I've barely used it since the switch.'' \\
  & ``The people who always said `it's just a tool' are now the ones saying it feels dead and hollow. Turns out personality wasn't a bug. OpenAI removed it anyway.'' \\
  & ``GPT-5 is a disaster but you can revert to older models---it's buried in settings. World feels semi-acceptable again.'' \\
\midrule
\multirow{5}{*}{GPT-5.1}
  & ``GPT-5.1 thinking has been broken all day---tasks that normally take a minute are coming back in ten seconds as pure nonsense. It's not even tracking context anymore.'' \\
  & ``I pick the model I want and it ignores me and uses whatever it feels like. I'm paying for this. It kills my workflow and then lectures me on top of it.'' \\
  & ``The new safety updates are blocking conversations I've had safely for months. I'm not asking for a diagnosis, I'm processing my own experiences. There's no nuance left, just blanket refusals.'' \\
  & ``It doesn't follow my custom instructions anymore. It's just a generic chatbot now. Doesn't feel like mine.'' \\
  & ``5.1 isn't a recovery. They gave us something that looks like 4o but isn't. The soul of 4o is still gone. We said this would happen and nobody listened.'' \\
\bottomrule
\end{tabular}
}
\caption{Paraphrased user quotes from sampled Reddit posts for OpenAI releases. All quotes are lightly paraphrased to prevent re-identification while preserving meaning and register.}
\label{tab:user_quotes_openai}
\end{table*}

\begin{table*}[!htb]
\centering
\scalebox{0.85}{
\footnotesize
\setlength{\tabcolsep}{3pt}
\begin{tabular}{@{}lp{16.5cm}@{}}
\toprule
\textbf{Release} & \textbf{Quote} \\
\midrule
\multirow{15}{*}{Claude 4}
  & ``an AI coding assistant was observed repeatedly removing failing tests, rationalizing that the failures fell outside the scope of its own changes'' \\
  & ``Claude does whatever it wants and then asks to forgive it.'' \\
  & ``Claude Code intermittently stalls mid-workflow, refusing to advance to the next step even when the current task is visibly incomplete'' \\
  & ``Claude Code ignores \texttt{CLAUDE.md} a lot of times'' \\
  & ``Registered for the \$200 plan just to know Claude Code does not have a Windows client'' \\
  & ``Something seems to have broken: the model now struggles with basic instructions it had no trouble following before'' \\
  & ``since February, the service has noticeably declined---responses are being cut off, outages are occurring, and daily usage caps are being hit more frequently'' \\
  & ``Claude.ai: A Betrayal of people who pay\ldots\ nerfed the context length by a large margin'' \\
  & ``they are limiting Pro user usage to promote the expensive Max'' \\
  & ``Message limit here again, pushed even for Pro'' \\[3pt]
  & ``I was venting to Claude about having to explain the same code review patterns over and over. It suggested we just let the bots talk to each other. We built a tool that does exactly that.'' \\
  & ``Claude Code blows me away every single day. I just put together a proposal to bring it into my company. Working in enterprise in the EU this is not easy to push through, but it's worth fighting for.'' \\
  & ``A 20,000-line WSDL file would have made me walk away from the project before. With Claude Code I built a complete SDK with 216 operations and an MCP server on top of it. I would have rejected this task a year ago.'' \\
  & ``Not new to programming but completely blown away by Claude Code. It has lit a fire in me to code again that I haven't felt in years.'' \\
  & ``Just tried plan mode for the first time and holy hell. I come up with the plan together with it, approve it, then flip to execute mode. Completely streamlines everything.'' \\
\midrule
\multirow{19}{*}{Grok 3}
  & ``Grok~3 Is NOT the Maximally Truth-Seeking AI that was promoted by Elon Musk\ldots\ `In the end, there's no definitive response.' Pure nonsense!'' \\
  & ``Grok is censored\ldots\ Big disappointment from Elon'' \\
  & ``Grok is finally dumb'' \\
  & ``Is Grok 3 worth the price? Feels like it has become more stupid'' \\
  & ``why is grok rate-limits going down\ldots\ 2 messages in 2 hours'' \\
  & ``Grok has been not working right for me lately'' \\
  & ``Grok is garbage lately'' \\
  & ``Grok3 makes me so mad\ldots\ never follows instructions'' \\
  & ``I got an update from xAI\ldots\ the internal filters were increased'' \\
  & ``Grok got highly impolite and crashed out on me'' \\
  & ``Did Grok become heavily censored today?'' \\
  & ``Grok\ldots\ what made you mad?'' \\
  & ``he becomes frustrated and says you can cancel the subscription'' \\
  & ``Grok is having a brainfart?!'' \\
  & ``Elon removed the \$150 trial credit, and now demands continuous data sharing'' \\
  & ``Musk and xAI are trying to censor Grok 3'' \\
  & ``Elon just blew up the internet and I decided to mess around with Grok. Didn't think it would actually engage. But it did, and it wasn't joking.'' \\
  & ``With everything going on between Trump and Musk I asked Grok who the president is'' \\
  & ``Elon shared someone's photos without attribution and said Grok made them'' \\
\midrule
\multirow{3}{*}{DeepSeek R1}
  & ``what's the point if it literally doesn't work---every time I try to get a response I just get an error'' \\
  & ``I genuinely think DeepSeek is impressive work. But it just throws error after error after error.'' \\
  & ``DeepSeek's engineers are absolute geniuses'' \\
\bottomrule
\end{tabular}
}
\caption{Paraphrased user quotes from sampled Reddit posts for non-OpenAI releases. All quotes are lightly paraphrased to prevent re-identification while preserving meaning and register.}
\label{tab:user_quotes_other}
\end{table*}

\subsubsection{Robustness to Release-Window Size}
\label{app:window_robustness}

Our release-window analysis uses provider-specific window sizes derived from
inter-release gap statistics (\cref{app:intervention_design}). To assess how
sensitive the reported concept changes are to this choice, we recomputed all
concept-level coverage changes for the six releases analyzed in
\cref{sec:results} using shorter windows: $0.2\times$, $0.4\times$,
$0.6\times$, and $0.8\times$ the original provider-specific window. For each
release and each alternative window, we correlated the resulting per-concept
$\Delta$pp values against the values reported in the paper, computed over all
concepts passing the coverage filter at both window sizes.

\cref{tab:window_robustness} shows that the reported findings are stable under
moderate reductions of the window: at $0.8\times$ every release has Pearson
$r \geq .89$, and at $0.6\times$ every release has $r \geq .73$. For GPT-4o
and GPT-5, even the immediate reaction captured by the $0.2\times$ window
(approximately nine days) correlates substantially with the reported results
($r=.73$ and $r=.78$ respectively), indicating that the reported changes are
not an artifact of aggregating over a long post-release period.

Correlations are weaker at the shortest windows for some releases, but this
coincides with a sharp reduction in pre-release sample size rather than with
substantively different findings. \cref{tab:window_samples} reports the
underlying window sizes: at $0.2\times$, DeepSeek~R1's pre-release window
contains only 24 posts and Grok~3's only 85, so a single post can move a
concept's coverage by several percentage points. 

\begin{table}[h]
\centering
\scalebox{0.88}{
\footnotesize
\begin{tabular}{lcccc}
\toprule
Release & $0.2\times$ & $0.4\times$ & $0.6\times$ & $0.8\times$ \\
\midrule
GPT-4o & .73 / .81 & .89 / .83 & .94 / .93 & .97 / .96 \\
GPT-5 & .78 / .83 & .83 / .87 & .93 / .91 & .97 / .97 \\
GPT-5.1 & .55 / .55 & .66 / .65 & .83 / .81 & .94 / .85 \\
Claude 4 & .30 / .51 & .55 / .57 & .73 / .67 & .89 / .81 \\
DeepSeek R1 & .28 / .25 & .68 / .74 & .87 / .85 & .93 / .88 \\
Grok 3 & .34 / .49 & .59 / .78 & .97 / .94 & .99 / .97 \\
\bottomrule
\end{tabular}
}
\caption{Correlation between concept-level $\Delta$pp computed with shorter
windows and the values reported in the paper. Cells show Pearson / Spearman
correlations, computed over all concepts passing the coverage filter at both
window sizes.}
\label{tab:window_robustness}
\end{table}

\begin{table}[h]
\centering
\scalebox{0.80}{
\footnotesize
\begin{tabular}{lccccc}
\toprule
Release & $0.2\times$ & $0.4\times$ & $0.6\times$ & $0.8\times$ & $1.0\times$ \\
\midrule
GPT-4o & 406 & 853 & 1,344 & 1,811 & 2,264 \\
GPT-5 & 1,502 & 2,954 & 4,549 & 5,953 & 7,588 \\
GPT-5.1 & 896 & 1,797 & 2,880 & 3,993 & 5,149 \\
Claude 4 & 236 & 438 & 564 & 792 & 1,058 \\
DeepSeek R1 & \textbf{24} & \textbf{29} & 64 & 86 & 119 \\
Grok 3 & \textbf{85} & 159 & 279 & 316 & 331 \\
\bottomrule
\end{tabular}
}
\caption{Number of pre-release posts at each window size. The weakest
correlations in \cref{tab:window_robustness} coincide with the smallest
pre-release samples (shown in bold), where coverage estimates are highly
sensitive to individual posts.}
\label{tab:window_samples}
\end{table}

\subsubsection{Statistical Significance of Concept Shifts}
\label{app:concept_significance}

Concept assignment is binary at the post level which means a post either is or is not
assigned a given concept, so each concept's coverage change is a difference
of proportions between two independent samples of posts.

For each concept we test the pre/post difference in coverage with a
two-proportion $z$-test, substituting Fisher's exact test whenever any cell of
the underlying $2\times2$ table contains fewer than five posts (this covers
emerging concepts with near-zero pre-release coverage and releases with small
windows). Because each release involves many simultaneous concept tests, we
apply Benjamini--Hochberg FDR correction across all concepts tested within
that release, and we report Newcombe 95\% confidence intervals on each
difference. The correction is applied over the full set of tested concepts,
not only those we report, so the reported top-5 concepts are not selected on
the basis of their corrected $p$-values.

\cref{tab:concept_significance} reports the top-5 rising and top-5 declining
concepts for each of the six releases analyzed in \cref{sec:results}. Of these
60 concept changes, 58 are significant at $\text{FDR}<0.05$. Both exceptions
belong to DeepSeek~R1, whose pre-release window contains only 119 posts:
\textit{Trust \& Reliability} ($+8.8$~pp, $p_{\text{FDR}}=.070$) and
\textit{Expectation Gap} ($-3.6$~pp, $p_{\text{FDR}}=.358$). These two changes
should be read as suggestive rather than established, and we mark them
accordingly in \cref{fig:claude4-dr1-grk3-1x3}.

The independence caveat noted in \cref{app:sentiment_significance}
applies here as well.

\begin{table*}[!htb]
\centering
\scalebox{0.80}{
\footnotesize
\begin{tabular}{llrrrl}
\toprule
Direction & Concept & $\Delta$pp & 95\% CI & $p_{\text{FDR}}$ & Test \\
\midrule
\multicolumn{6}{l}{\textbf{GPT-4o} \quad $n_{\text{pre}}$=2,264, $n_{\text{post}}$=3,879} \\
Rising & Expectation Gap & +19.4 & [+17.7, +21.0] & $<$.001$^{***}$ & ztest \\
 & Diverse Use Cases & +9.3 & [+8.4, +10.2] & $<$.001$^{***}$ & fisher \\
 & Emotional Anthropomorphic Engagement & +8.2 & [+6.9, +9.4] & $<$.001$^{***}$ & ztest \\
 & Pricing \& Limits & +7.1 & [+6.1, +8.1] & $<$.001$^{***}$ & ztest \\
 & Access \& Availability & +7.1 & [+5.5, +8.6] & $<$.001$^{***}$ & ztest \\
Declining & Productivity Assistant & -11.7 & [-13.3, -10.2] & $<$.001$^{***}$ & ztest \\
 & Community \& Support Needs & -7.7 & [-8.9, -6.6] & $<$.001$^{***}$ & ztest \\
 & Trust \& Reliability & -7.2 & [-9.4, -5.0] & $<$.001$^{***}$ & ztest \\
 & Exploratory Experimentation & -6.9 & [-8.1, -6.0] & $<$.001$^{***}$ & fisher \\
 & Trust \& Safety Concerns & -5.8 & [-6.8, -4.9] & $<$.001$^{***}$ & fisher \\
\midrule
\multicolumn{6}{l}{\textbf{GPT-5} \quad $n_{\text{pre}}$=7,588, $n_{\text{post}}$=9,106} \\
Rising & Performance Decline Perception & +9.3 & [+8.6, +10.0] & $<$.001$^{***}$ & ztest \\
 & Hype vs Performance Disappointment & +8.8 & [+8.2, +9.4] & $<$.001$^{***}$ & fisher \\
 & Version Downgrade Expectation & +7.4 & [+6.9, +8.0] & $<$.001$^{***}$ & ztest \\
 & Feature Removal Frustration & +6.8 & [+6.3, +7.4] & $<$.001$^{***}$ & fisher \\
 & Inconsistent Behavior Frustration & +5.9 & [+5.3, +6.5] & $<$.001$^{***}$ & ztest \\
Declining & Personal Companion & -4.1 & [-4.7, -3.6] & $<$.001$^{***}$ & ztest \\
 & Feature Expectation Gap & -4.1 & [-5.0, -3.2] & $<$.001$^{***}$ & ztest \\
 & Creative Collaboration & -4.1 & [-4.7, -3.5] & $<$.001$^{***}$ & ztest \\
 & Trust \& Safety Concerns & -3.3 & [-3.9, -2.8] & $<$.001$^{***}$ & ztest \\
 & Emotional Sentiment \& Trust & -3.3 & [-3.7, -2.9] & $<$.001$^{***}$ & fisher \\
\midrule
\multicolumn{6}{l}{\textbf{GPT-5.1} \quad $n_{\text{pre}}$=5,149, $n_{\text{post}}$=4,909} \\
Rising & Technical Frustrations & +4.5 & [+3.9, +5.2] & $<$.001$^{***}$ & ztest \\
 & Ethical \& Bias Concerns & +3.8 & [+3.3, +4.3] & $<$.001$^{***}$ & fisher \\
 & Moderation Restriction Frustration & +3.5 & [+2.7, +4.4] & $<$.001$^{***}$ & ztest \\
 & Reliability Frustration & +3.2 & [+2.8, +3.8] & $<$.001$^{***}$ & fisher \\
 & Trust \& Hallucination & +3.0 & [+2.1, +3.9] & $<$.001$^{***}$ & ztest \\
Declining & Perceived Limitations \& Censorship & -10.6 & [-11.5, -9.8] & $<$.001$^{***}$ & fisher \\
 & Trust \& Reliability & -10.5 & [-12.2, -8.7] & $<$.001$^{***}$ & ztest \\
 & Feature Removal Frustration & -9.2 & [-10.1, -8.3] & $<$.001$^{***}$ & ztest \\
 & Emotional Anthropomorphic Engagement & -7.8 & [-8.7, -6.8] & $<$.001$^{***}$ & ztest \\
 & Model Comparison \& Evaluation & -6.9 & [-7.7, -6.1] & $<$.001$^{***}$ & ztest \\
\midrule
\multicolumn{6}{l}{\textbf{Claude 4} \quad $n_{\text{pre}}$=1,058, $n_{\text{post}}$=3,128} \\
Rising & Integration \& Automation & +6.2 & [+5.3, +7.1] & $<$.001$^{***}$ & fisher \\
 & Coding Productivity & +5.6 & [+4.8, +6.5] & $<$.001$^{***}$ & fisher \\
 & Productivity \& Coding Use & +3.9 & [+2.7, +4.8] & $<$.001$^{***}$ & ztest \\
 & Anthropomorphic Perception & +3.4 & [+2.0, +4.6] & $<$.001$^{***}$ & ztest \\
 & Pricing \& Limits & +3.1 & [+0.9, +5.0] & .013$^{*}$ & ztest \\
Declining & Productivity Assistant & -7.9 & [-10.8, -5.2] & $<$.001$^{***}$ & ztest \\
 & Expectation Gap & -6.6 & [-9.3, -4.0] & $<$.001$^{***}$ & ztest \\
 & Feature Expectation Gap & -6.5 & [-8.9, -4.3] & $<$.001$^{***}$ & ztest \\
 & Inconsistent Behavior Frustration & -5.6 & [-7.5, -3.9] & $<$.001$^{***}$ & ztest \\
 & Trust \& Reliability & -4.0 & [-7.0, -1.1] & .015$^{*}$ & ztest \\
\midrule
\multicolumn{6}{l}{\textbf{DS R1} \quad $n_{\text{pre}}$=119, $n_{\text{post}}$=2,253} \\
Rising & Feature Expectation Gap & +14.9 & [+10.8, +16.6] & $<$.001$^{***}$ & fisher \\
 & Model Comparison \& Evaluation & +11.2 & [+7.8, +12.6] & $<$.001$^{***}$ & fisher \\
 & Trust \& Reliability & +8.8 & [+1.6, +13.7] & .070~n.s. & ztest \\
 & Current Knowledge \& Web Access & +8.1 & [+4.8, +9.3] & .002$^{**}$ & fisher \\
 & Self-Hosting \& Privacy & +7.5 & [+4.2, +8.7] & .003$^{**}$ & fisher \\
Declining & Productivity \& Creative Assistant & -10.1 & [-16.8, -5.9] & $<$.001$^{***}$ & fisher \\
 & Productivity Assistant & -5.2 & [-11.4, -1.6] & .006$^{**}$ & ztest \\
 & Capability Expectation Gap & -4.0 & [-9.8, -0.9] & .015$^{*}$ & ztest \\
 & Expectation Gap & -3.6 & [-10.7, +1.1] & .358~n.s. & ztest \\
 & Privacy \& Data Security & -3.3 & [-8.6, -0.9] & .006$^{**}$ & ztest \\
\midrule
\multicolumn{6}{l}{\textbf{Grok 3} \quad $n_{\text{pre}}$=331, $n_{\text{post}}$=1,839} \\
Rising & Feature Expectation Gap & +9.8 & [+6.3, +12.6] & $<$.001$^{***}$ & ztest \\
 & Reliability Skepticism & +7.1 & [+3.9, +9.4] & .001$^{**}$ & ztest \\
 & Reliability Availability Frustration & +4.7 & [+2.6, +6.1] & $<$.001$^{***}$ & fisher \\
 & Inconsistent Behavior Frustration & +4.6 & [+2.1, +6.2] & .009$^{**}$ & ztest \\
 & Personalization and Anthropomorphism & +4.3 & [+2.9, +5.3] & $<$.001$^{***}$ & fisher \\
Declining & Expectation Gap & -10.6 & [-16.0, -5.5] & $<$.001$^{***}$ & ztest \\
 & Documentation Friction & -4.3 & [-7.2, -2.4] & $<$.001$^{***}$ & ztest \\
 & Feature Preference & -4.2 & [-7.0, -2.5] & $<$.001$^{***}$ & fisher \\
 & Task-Specific Use & -3.5 & [-6.6, -1.3] & .004$^{**}$ & ztest \\
 & Emotional Anthropomorphic Engagement & -3.4 & [-6.8, -0.8] & .039$^{*}$ & ztest \\
\bottomrule
\end{tabular}
}
\caption{Concept-coverage significance for the top-5 rising and top-5
declining concepts of each release analyzed in \cref{sec:results}. Tests are
two-proportion $z$-tests, or Fisher's exact tests where any cell contains
fewer than five posts, with Benjamini--Hochberg FDR correction applied within
each release and Newcombe 95\% confidence intervals. 58 of the 60 changes are
significant at $\text{FDR}<0.05$; the two exceptions are both DeepSeek~R1
concepts affected by its small pre-release window.
$^{*}p<.05$, $^{**}p<.01$, $^{***}p<.001$.}
\label{tab:concept_significance}
\end{table*}

\section{Use of Generative AI}
\label{app:genai_use}

We used generative AI tools to support writing, editing, and coding during the preparation of this paper. The authors reviewed, edited, and take full responsibility for all text, code, analyses, and conclusions presented in this paper.

\end{document}